\pdfoutput=1
\documentclass[11pt]{article}
\PassOptionsToPackage{table,dvipsnames}{xcolor}

\usepackage[final]{acl}
\usepackage{times}
\usepackage{titlesec}
\usepackage{latexsym}
\usepackage{soul}
\usepackage{multirow}
\usepackage{siunitx}
\usepackage{amsfonts}
\usepackage{arydshln} 

\usepackage[T1]{fontenc}

\usepackage[utf8]{inputenc}

\usepackage{microtype}

\usepackage{inconsolata}

\usepackage{booktabs}
\usepackage{algorithm}
\usepackage{algpseudocode}
\usepackage{amsmath}
\usepackage{pifont}
\usepackage{tcolorbox}
\usepackage{hyperref}
\usepackage{graphicx}
\usepackage{tabularx}
\usepackage{verbatim}
\usepackage{float}
\usepackage{colortbl}   
\usepackage{paralist}

\usepackage{tikz}

\title{NLKI: A \textit{lightweight} Natural Language Knowledge Integration Framework for Improving Small VLMs in Commonsense VQA Tasks}

\author{
  Aritra Dutta \\
  IIT Kharagpur \\
  \And
  Swapnanil Mukherjee \\
  Ashoka University \\
  \And
  Deepanway Ghosal \\
  Independent Researcher \thanks{Now at Google} \\
  \And
  Somak Aditya \\
  IIT Kharagpur \\
}

\begin{document}
\maketitle
\begin{abstract}
Commonsense visual–question answering often hinges on knowledge that is missing from the image or the question. Small vision-language models (sVLMs) such as ViLT, VisualBERT and FLAVA therefore lag behind their larger generative counterparts. To study the effect of careful commonsense knowledge integration on sVLMs, we present an end-to-end framework (NLKI) that (\textit{i}) retrieves natural language facts, (\textit{ii}) prompts an LLM to craft natural language explanations, and (\textit{iii}) feeds both signals to sVLMs respectively across two commonsense VQA datasets (CRIC, AOKVQA) and a visual-entailment dataset (e-SNLI-VE). Facts retrieved using a fine-tuned ColBERTv2 and an object information-enriched prompt yield explanations that largely cut down hallucinations, while lifting the end-to-end answer accuracy by up to 7\% (across 3 datasets), making FLAVA and other models in NLKI match or exceed medium-sized VLMs such as Qwen-2 VL-$2B$ and SmolVLM-$2.5B$. As these benchmarks contain 10–25\% label noise, additional finetuning using noise-robust losses (such as symmetric cross entropy and generalised cross entropy) adds another $2.5\%$ in CRIC, and $5.5\%$ in AOKVQA.  
Our findings expose when LLM-based commonsense knowledge beats retrieval from commonsense knowledge bases, how noise-aware training stabilises small models in the context of external knowledge augmentation, and why parameter-efficient commonsense reasoning is now within reach for $250M$ models. \footnote{We release our code, checkpoints, and data at: \href{https://github.com/beingdutta/NLKI-Lightweight-Natural-Language-Knowledge-Integration-Framework}{https://github.com/beingdutta/NLKI-Lightweight-Natural-Language-Knowledge-Integration-Framework}}

\end{abstract}

\section{Introduction}
\begin{figure}[t]
\centering
\includegraphics[width=\linewidth]{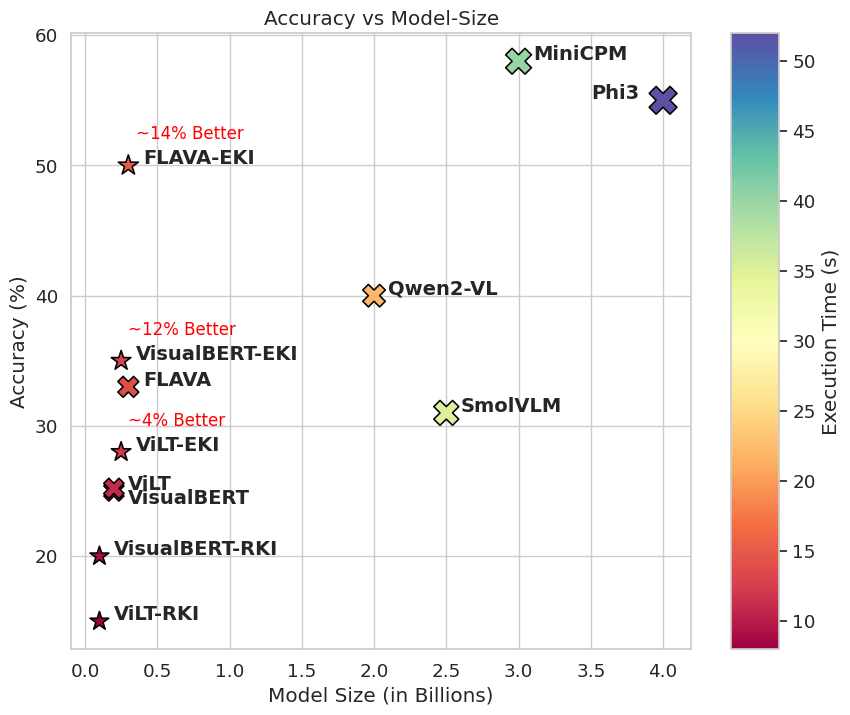} 
\caption{Comparison of accuracy and model size for various small Vision-Language Models (sVLMs) AOKVQA (Val). $EKI$: Type-5 Explanation Knowledge Integrated models, as defined in \ref{tab:main_accuracy_table}, while $RKI$: Retrieved Knowledge Integrated models. The colour gradient represents execution time, with red indicating longer durations and blue indicating shorter ones.}
\label{fig:accuracy-vs-size}
\end{figure}
Early Visual Question Answering (VQA) work already recognised that images and questions alone are often insufficient and called for the use of external knowledge sources \cite{kbvqawang2015explicit,aditya2018explicit,aditya2019integrating}. Factoid VQA now routinely benefits from graph‐ or web‐scale retrieval \cite{fvqa2018,okvqaMarino_2019_CVPR,chen2024knowledge}, yet the commonsense variant remains under-explored, especially for \emph{small} vision–language models (sVLMs) such as \textsc{ViLT}, \textsc{VisualBERT} and \textsc{FLAVA}. Recent attempts either bypass pre-trained small VLMs \cite{ye2023improvingdance} or limit themselves to factual QA \cite{lin-byrne-2022-retrieval-ravqa,yang-etal-2023-vilm}. A big challenge in commonsense knowledge retrieval is a lack of a unified source -- no single knowledge graph, corpus, or model covers everyday physics, social conventions, and object affordances. Large language models (LLMs) have emerged as promising but noisy reservoirs of such knowledge \cite{ghosal-etal-2023-language}. These challenges motivate our four research questions: \textbf{RQ1)} \emph{to what extent can commonsense facts retrieved from knowledge bases and LLM-generated explanations help small VLMs?}; And \textbf{RQ2)} \emph{is one better suited than the other for commonsense reasoning in VQA systems?}; \textbf{RQ3)} \emph{can noise-robust losses mitigate the pervasive label noise in the context of external knowledge augmentation?}; and to draw relevance to recent progress in lightweight generative AI, we analyse \textbf{RQ4)} \emph{how do medium-size generative VLMs ($\leq 4B$) fare in commonsense tasks, in comparison to sVLMs (with or without knowledge integration)?}

To probe \textbf{RQ1–2}, we test the three sVLMs above ($\leq 240 M$ parameters) on CRIC and AOKVQA \cite{cric,aokvqa} along with a visual entailment dataset, e-SNLI-VE \cite{snlive}. Using ground-truth explanations (along with the image and question) lifts accuracy by up to \emph{10 to 12\%}, signalling the necessity and scope for utilising text-based knowledge. We therefore contrast (\emph{i}) ColBERTv2-retrieved facts from commonsense Knowledge Bases (KBs) and (\emph{ii}) free-form textual explanation produced by Llama-3-8B from dense/region captions, list of objects, and retrieved facts. A richer visual context helps curb hallucination, while simple architectural tweaks temper the effect of occasional noisy snippets. Because label noise dominates these datasets, \textbf{RQ3} tests whether Generalised Cross-Entropy (GCE) \cite{generalizedcrossentropy} and Symmetric Cross-Entropy (SCE) \cite{symmetriccrossentropy} preserve the gains and further boost the performance of knowledge integration, while mitigating the effects of label noise. Finally, addressing \textbf{RQ4}, we evaluate Qwen-2, Phi-3-Vision, MiniCPM and SmolVLM ($\leq 4 B$) and show that they too lack commonsense without explicit knowledge integration, underscoring the urgency of lightweight, knowledge-aware methods. Overall, we make the following contributions: 

\begin{compactitem}
\item \textbf{Plug-and-Play Knowledge Integration (NLKI)}: We introduce \textbf{N}atural-\textbf{L}anguage \textbf{K}nowledge \textbf{I}ntegration (NLKI) -- a plug-and-play framework that combines a retriever, an LLM explainer, and a lightweight reader with any sub-$240M$ VLM, so each module can be analysed and improved independently to improve the overall accuracy of small VLMs in commonsense reasoning tasks.
\item \textbf{Dense retrieval benchmark}: Compared to various state-of-the-art retrievers, we showed that finetuning ColBERTv2 using contrastive learning delivers the highest recall in retrieving the most relevant commonsense facts for CRIC, AOKVQA, and e-SNLI-VE queries.
\item \textbf{Accurate LLM Explanation using Context and Knowledge-Enriched Prompting}: We show that adding object and region information of an image (through dense/region captions generated from Florence-2-large) along with the retrieved facts to the Llama-3.1-8B prompt sharply reduces noise and hallucinated details in the generated explanation, compared to a caption-only context. 
\item \textbf{Knowledge \& noise-robustness gains}: LLM-generated explanations coupled with Symmetric Cross Entropy (SCE) or Generalised Cross-Entropy (GCE) preserve most of that gain under heavy label noise scenarios, raising AOKVQA accuracy by an average of $13.6\%$ and CRIC, e-SNLI-VE by $2-4\%$ across three architectures;
\item \textbf{Scale comparison}: Evaluations of Qwen-2, Phi-3-Vision, MiniCPM, and SmolVLM ($\leq 4 B$) show these larger models still lack commonsense, while NLKI-equipped small VLMs ($\leq 240 M$) can match or outperform them in some cases.
\end{compactitem}

\section{Related Work}
Knowledge-based VQA systems pose a challenging task, as they need the models to comprehend visual and textual data while utilising external knowledge to derive correct answers. For instance, as depicted in Fig.~\ref{fig:pt_vs_ft_colbert}, \textit{``Is there a place that is blue and liquid?''}. The correct result ``Ocean'' has to be deduced using the visual information, common knowledge that it is ``liquid'' and ``blue''.
Symbolic structured knowledge is helpful in this context, for instance, existing literature has used knowledge graphs to retrieve facts, using knowledge graph embeddings and formal language queries \cite{fvqa2018,ZHENG2021108153,chen2021zeroshot}. Beyond KGs, logic-based formalisms have also been used to combine knowledge with reasoning, e.g., Probabilistic Soft Logic for visual puzzle solving in \cite{aditya2018combining}. For unstructured knowledge, \citet{karpukhin2020dense} showed that the Dense-Passage-Retrieval (DPR) technique performs the best by encoding both the question and the knowledge sentences to text embedding using BERT. This method uses cosine similarity (or the dot product) between vectors to retrieve relevant knowledge text.

\paragraph{Retrieval-Augmentation for sVLM.} 
Retrieval-augmented learning is widely used in language modelling and is increasingly explored in vision-language and multimodal tasks. Recent studies apply retrieval-based augmentation in several ways:
(i) Image-based retrieval: \citet{rao2024raven} retrieved image-caption pairs using a CLIP-based \cite{radford2021learning} scoring technique, processed by an OFA model ($182M$ parameters) for text generation. \citet{wang2023visuallyaugmented} integrated similar images into vision-language attention layers in a transformer decoder. \citet{rao2023retrieval} retrieved knowledge graphs (KGs), encoded them with a graph encoder, and used them in a BERT + ViT-B vision-language model. These approaches show promising results by enhancing VQA and image captioning through retrieval mechanisms.\\\noindent
(ii) Multimodal retrieval: \citet{hu2023reveal} and \citet{yasunaga2023retrieval} proposed multimodal retrieval techniques compatible with language modeling, with \citet{hu2023reveal} using T5 + ViT ($400M$–$2.1B$ parameters). \citet{caffagni2024wiki} proposed a two-stage retrieval: CLIP retrieves documents, and Contriever extracts passages for LLaVA VLMs (7B+ parameters) in visual QA. While similar, our approach incorporates additional textual cues, expands retrieval beyond Wikipedia, and uses a smaller VLM.

\section{Problem Formulation}
\vspace{-0.5em}
In Visual Question Answering (VQA), some questions require knowledge beyond what is directly visible in an image. For example, identifying a giraffe as an \textit{"even-toed ungulate"} or recognising a sofa as \textit{"furniture used for sitting"} requires commonsense reasoning. Ambiguities in commonsense datasets further challenge models, as multiple correct answers may exist, making implicit knowledge crucial for accurate predictions. Addressing these gaps is essential for improving the VQA system's contextual understanding.
Here, we lay down the task more formally. The input for our task comprises an image ($I$) and a natural language question ($Q$), and the objective is to answer the question while utilising a set of implicitly required commonsense knowledge facts (s) ($K'$). We utilise the ground-truth answer ($A$) and the ground-truth explanation/knowledge ($f^*$), when available, as supervision. 

\section{Knowledge-Augmented Visual QA framework}

Here, we explore two types of knowledge sources: traditional commonsense knowledge corpora (textual versions) and pre-trained Large Language Models. Following \citet{chen2024knowledge}, we can represent a retrieval or knowledge-augmented VQA system probabilistically as follows:
\begin{equation*}
    p(A|I,Q,\mathcal{K}) = q_\Theta(\mathcal{F}_{ret}|Q,I,\mathcal{K})p_\Phi(A|Q,I,\mathcal{F}_{ret}),
\end{equation*}
where, $q_\Theta(\cdot)$ is termed as a retriever, which is tuned to retrieve top $k$ relevant facts ($\mathcal{F}_{ret}$), and $p_\Phi(\cdot)$ is termed as the reader, which utilizes the question, image, and retrieved facts to predict the answer. To generalise to LLM-generated explanations, we can think of the reader as a model that outputs a paragraph relevant to the question. We will utilise the following notations wherever required:
\begin{equation*}
    p(A|I,Q,\mathcal{M}) = q_\Theta(\mathcal{E}_{ret}|Q,I,\mathcal{M})p_\Phi(A|Q,I,\mathcal{E}_{ret}),
\end{equation*}
where $\mathcal{M}$ may stand for a knowledge base or an LLM. $\mathcal{E}_{ret}$ may stand for a set of retrieved facts and LLM-generated explanation(s).

\subsection{Integrating Knowledge from Commonsense Corpus} 

\paragraph{Commonsense Corpus.}
Commonsense knowledge comprises basic facts and anecdotes about everyday objects, actions, and events, enabling humans to make inferences \cite{li2022survey}. Since Transformer models reason well with text-based commonsense knowledge \cite{ruletakerclark20}, we adopt Natural Language Knowledge Integration (NLKI), which is more flexible than structured knowledge graphs. \citet{raco} introduced a $20$M-fact natural language commonsense corpus from multiple human-annotated sources and web data. Based on initial ablations (Appendix Tables \ref{tab:cric_omcs_vs_20M_corpus} and \ref{tab:esnli_omcs_vs_20M_corpus}), we selected a subset: Open Mind Commonsense (OMCS) \cite{omcshavasi2010} for its higher similarity scores with ground-truth explanations and its compact size ($\sim$1.5M facts) describing everyday events and objects.

\paragraph{Commonsense Knowledge Retrieval.} Given an image-question pair, the goal is to retrieve $k$ relevant commonsense facts ($f_{i\leq k}$) from the commonsense corpus ($\mathcal{K}$) to fill in missing contextual knowledge. Ideally, retrieval involves abductive reasoning to infer necessary deductions. We primarily explore text-based dense retrieval using SBERT \cite{sbert}, ColBERTv2 \cite{colbert}, and  Stella selected from the MTEB leaderboard \cite{mteb}.

We represent the query in the following ways: 
\begin{compactitem}
    \item \textbf{Question}: We use the question as the query.
    \item \textbf{Caption + Question}: We generated image captions using BLIP-2 \cite{blip2}. We prepend the caption to the question to form the query.
    \item \textbf{Objects + Question}: We use a pre-trained YOLOv8 \cite{yolo} to detect objects from the image. We use the list of detected objects along with the question as the query.
    \item \textbf{Scene Graph Text + Question}: We generated scene graph from images using Relation Transformers \cite{cong2023reltr}. We prepend the sequence of scene graph triplets to the question as the query.
\end{compactitem}
We use SBERT \cite{sbert} to embed the commonsense corpus ($\mathcal{K}$), indexing it with FAISS \cite{faiss}. Queries (e.g., question/hypothesis) are encoded using the same model, and Nearest Neighbour Search retrieves the top-$k$ closest facts. Since SBERT captures semantic similarity, we further employ ColBERTv2, optimised for retrieval. Improved results (Table \ref{tab:retrieval_scores}) led us to fine-tune ColBERTv2 for commonsense fact retrieval.

\begin{figure}[h!]
    \centering
    \includegraphics[width=1.0\columnwidth, height=6.5cm]{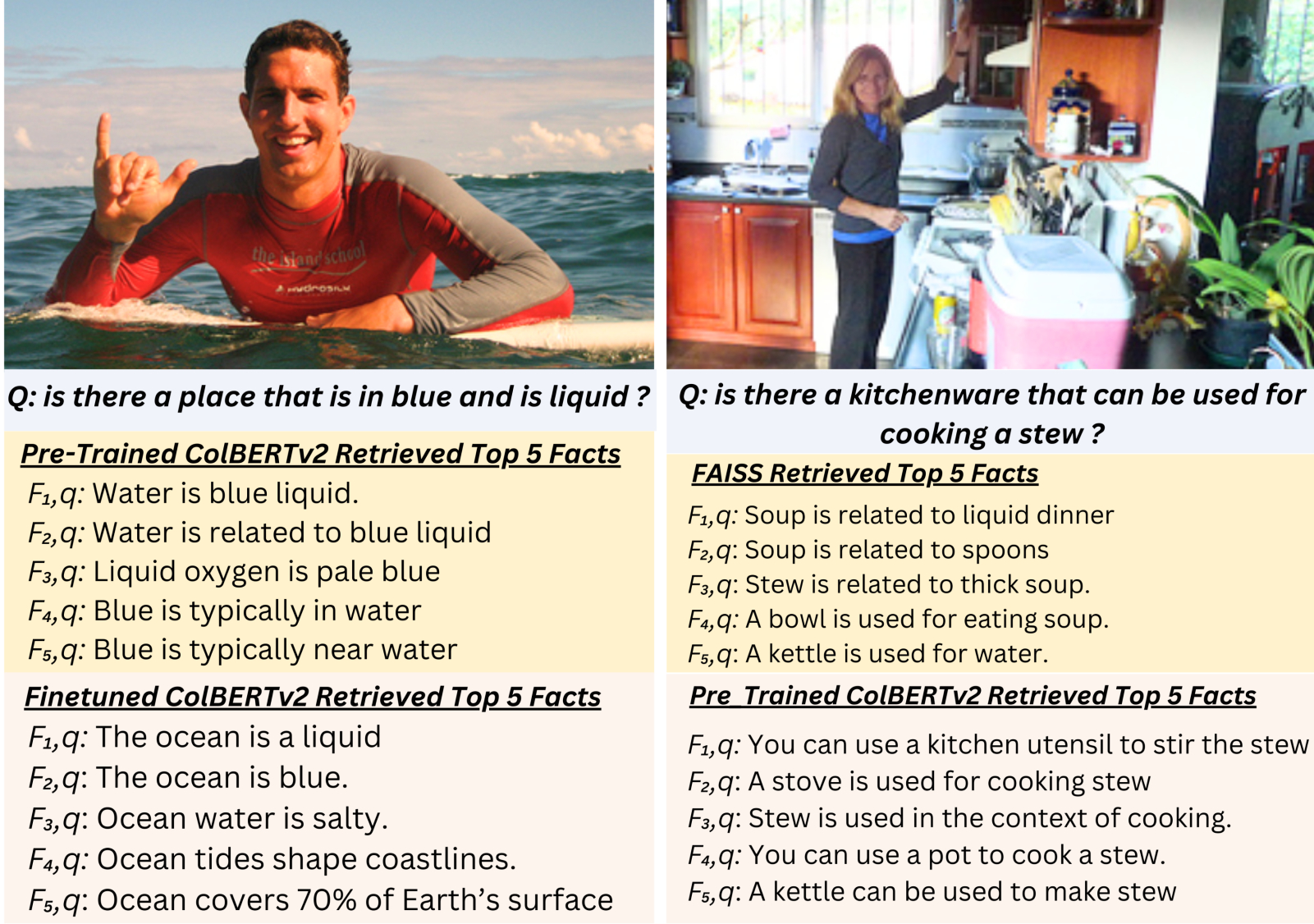} 
    \caption{Facts retrieved by pre-trained ColBERTv2 vs. Finetuned ColBERTv2 vs. FAISS. Facts retrieved using pre-trained ColBERTv2 are more semantically close and contextually relevant to the query compared to the FAISS similarity search over SBERT embeddings.}
    \label{fig:pt_vs_ft_colbert}
\end{figure}

\paragraph{Finetuning ColBERTv2.} As shown in Fig.~\ref{fig:pt_vs_ft_colbert}, pre-trained ColBERTv2 retrieves partially relevant facts since it is trained on generalist datasets (MS-MARCO, TREC-CAR). To improve relevance, we fine-tune it following \citet{colbert}, formatting the commonsense corpus into triples \(\langle q, d^+, d^- \rangle\) where $q$ is the query, $d^+$ is the ground-truth knowledge, and $d^-$ is a randomly sampled fact from the corpus (see Appendix \ref{sec:finetuning-colbert} for finetuning details).

\paragraph{Commonsense Knowledge Integration.}
The retriever-reader architectures can be trained in two ways: (\textit{i}) in an end-to-end fashion where the retriever and the VLM are finetuned together \cite{10.5555/3540261.3542249}, and (\textit{ii}) where they are finetuned independently of each other. For simplicity, we go with the latter approach to improve both independently, as this allows for a modular plug-and-play framework wherein the retriever/reader can be switched as per the specific task. And in the case of sub-optimal retrieval performance, the reader ($p_\Phi(\cdot)$) is expected to learn to reason with noisy added knowledge.

\subsection{Integrating LLM-generated Explanations}
\label{sec:explanationtypes}
Using the image $I$, we extract various pieces of information such as multiple types of captions, a list of objects, and a scene graph to textually represent the visual context, which is also sometimes paired with the retrieved facts (these can be perceived as parameters to the prompt). To define an oracle setting (\textbf{Type 0}), we also supply the ground-truth label (GT-L). Interestingly, Llama-3.1-8B produces high-quality explanations when provided with the ground-truth label. Manual analysis shows that such explanations are riddled with hallucinations when the captions do not capture enough question-relevant information or the label is noisy in the case of Type-0 (See Fig. \ref{fig:hallucinated_explanation_collection}). Therefore, we performed a detailed analysis of the effect of varying all of the above parameters in generating an explanation.  We provide the common instruction in Fig.~\ref{prompt}, as utilised for the instruction-finetuned LLMs (such as Llama-3.1-8B) to generate explanations\footnote{For Llama-3.1-8B we used the official implementation from Huggingface at \href{https://huggingface.co/meta-llama/Llama-3.1-8B}{\url{https://huggingface.co/meta-llama/Llama-3.1-8B}}}. The variants are the following: 
 \begin{compactitem}
 \item \textbf{Type-0}: TC + Q + $RF_{I,Q}$ + GT-L
     \item \textbf{Type-1}: TC + Q + $RF_{I,Q}$
     \item \textbf{Type-2}: DC + Q + $RF_{I,Q}$
     \item \textbf{Type-3}: RC + Q + $RF_{I,Q}$
     \item \textbf{Type-4}: RC + Q + O + $RF_{I,Q}$
     \item \textbf{Type-5}: DC + RC + Q + O + $RF_{I,Q}$
     \item \textbf{Type-6}: DC + RC + Q + O (Type 5 without knowledge facts).
     \item {\textbf{Type-7}: Florence Finetuned Explanations},
 \end{compactitem}
where $TC$: Traditional Image Caption, $DC$: Dense Caption, $RC$: Region Based Caption, $Q$: Question, $RF$: Retrieved Facts, $O$: Objects, GT-L: Ground Truth Label.

\paragraph{Integrating Explanations.} Since explanations are more targeted and concise, we explore baseline variations by simply prepending the explanation with the question for the small VLMs.

\section{Experimental Setup}
\subsection{Datasets}
In this work, we choose two popular VQA and one Natural Language Inference (NLI) datasets that require commonsense reasoning: CRIC \cite{cric}, AOKVQA \cite{aokvqa} and e-SNLI-VE \cite{snlive} due to their size, diversity of questions, and availability of ground truth explanations. In this work, we do not consider datasets that test (encyclopedic) knowledge about entities such as FVQA \cite{fvqa2018}, KB-VQA \cite{kbvqawang2015explicit}. More details about our choice of datasets are in Appendix \ref{sec:dataset-descriptions}.

\begin{table*}[!htb]
\small
\centering
\begin{tabular}{ccccccccc}
\toprule
\textbf{Type} & \textbf{Explanation} & \textbf{B-1} & \textbf{B-2} & \textbf{B-3} & \textbf{R-1} & \textbf{R-2} & \textbf{R-L} & \textbf{Cosine} \\ 
\midrule
\textbf{Type-0} & TC + Q + $F_{I,Q}$ + $y_{I,Q}$ & 18.60 & 11.41 & 8.32 & 36.01 & 12.88 & 32.08 & 53.21 \\
\textbf{Type-1} & C + Q + $F_{I,Q}$ & 24.65 & 16.64 & 12.81 & 40.00 & 17.75 & 36.80 & 54.75 \\
\midrule
\textbf{Type-2} & DC + Q + $F_{I,Q}$ & 27.02 & 18.85 & 14.84 & 42.25 & 20.26 & 38.86 & 58.15 \\
\textbf{Type-3} & RC + Q + $F_{I,Q}$ & 30.84 & 22.57 & 18.16 & 45.90 & 24.51 & 42.80 & 59.60 \\
\textbf{Type-4} & RC + O + Q + $F_{I,Q}$ & 30.46 & 22.30 & 18.01 & 46.01 & 24.35 & 42.44 & 60.08 \\
\textbf{Type-5} & DC + RC + O + Q + $F_{I,Q}$ & \textbf{31.08} & \textbf{22.77} & \textbf{18.41} & \textbf{46.30} & \textbf{24.63} & \textbf{42.78} & \textbf{60.17} \\
\midrule
\textbf{Type-6} & DC + RC + O + Q & 25.88 & 18.51 & 14.67 & 41.93 & 20.86 & 38.57 & 57.01 \\
\textbf{Type-7} & I + $<TP>$ & 7.20 & 2.97 & 1.87 & 11.76 & 00.75 & 10.49 & 22.39 \\
\bottomrule
\end{tabular}
\caption{Results from CRIC (AOKVQA in App. \ref{sec:explanation-scores-aokvqa}) comparing BLEU, ROUGE, and Cosine Scores (w.r.t the GT explanation) for various explanation generation mechanisms. We report the quality metrics of explanations based on BLEU-$k$ (\textbf{B-$k$}, $k\in\{1,2,3\}$), ROUGE-$k$ (\textbf{R-$k$}, $k\in\{1,2,L\}$), and Cosine similarity scores. $Q$: Question, $O$: Objects, $TC$: Traditional Image Caption, $F_{I,Q}$: Retrieved Facts, $DC$: Dense Caption, $RC$: Region Caption, $I$: Image and $TP$: Task Prompt. These scores can be compared with the retrieved facts score metrics \textbf{B-1}@5, \textbf{R-L}@5, \textbf{Cosine}@5 of Tab.~\ref{tab:retrieval_scores}}.
\label{tab:explanation_quality_scores}
\end{table*}

\subsection{Architectures \& Training}
\label{sec:architectures}
We choose three small-vision-Language models -- VisualBERT \cite{li2019visualbert}, ViLT \cite{vilt1}, FLAVA \cite{flava} -- spanning two broad modality fusion architectures, all under \emph{240M} parameter size, including one dual stream architecture, and two single stream architectures. We briefly describe each of these architectures for our VQA task in Appendix \ref{sec:models-used}. For finetuning, we use the standard cross-entropy loss objective (and extend this with noise-robust variants in \S\ref{sec:noise-robust}). We have tried multiple variants to integrate commonsense knowledge, such as \emph{majority voting} (details in Appendix \ref{majority}), \emph{string concatenation}, and their variants with noise-robust loss functions.

\section{Results}
\label{results}
\subsection{Analysis of Standalone Retrieval Performance}
We evaluate retrieval methods to identify the best retriever component of our \emph{NLKI pipeline}. We compare the relevance of retrieved facts (by each of the retrievers) with ground-truth explanations for each of the datasets. As per our studies specific to commonsense knowledge, FAISS, which relies on pre-trained SBERT embeddings, struggles with retrieving relevant \emph{missing} information. Fig.~\ref{fig:pt_vs_ft_colbert} shows how the facts retrieved by FAISS are conceptually close to the question but not relevant, whereas the facts retrieved by ColBERTv2 are exact. In contrast, ColBERTv2 retrieves more precise facts and significantly outperforms FAISS and other state-of-the-art retrievers like Stella-$400$M\footnote{We used the official HuggingFace implementation at \href{https://huggingface.co/NovaSearch/stella_en_400M_v5}{\url{https://huggingface.co/NovaSearch/stella\_en\_400M\_v5}}} across BLEU, ROUGE, and cosine similarity metrics (see Tab.~\ref{tab:retrieval_scores}).

\subsection{Capturing Visual Context for LLM-Generated Explanations}
Table~\ref{tab:explanation_quality_scores} compares prompt settings for explanation generation. Type-5, incorporating dense captions, region captions, retrieved facts, and objects, outperforms other variants, effectively capturing colour information and context. Our manual analysis of baselines suggested that sVLMs struggle with noisy, ambiguous and colour-based questions. As the region and dense captions capture colour information, Type-5 explanations prove to be the most effective and relevant (compared to generated explanations of other \textit{Types} and retrieved facts standalone). We further explored Smaller LLaMA $3.2$ models ($3$B \& $1$B) to generate explanations. Such explanations turned out to be noisier and less accurate with poorer visual context, making them unsuitable (Detailed results in Appendix~\ref{sec:llama-variants}).

\subsection{Baseline Performance}
To estimate the vanilla performance (without knowledge augmentation), we finetuned ViLT, VisualBERT, and FLAVA with the image, the question (hypothesis) and the answer (label) across all our datasets. We report the results in Table~\ref{tab:main_accuracy_table}. Manual analysis (see \S\ref{sec:noise-robust}) reveals that the sub-optimal result is due to 1) \emph{lack of day-to-day commonsense knowledge} and 2) \emph{noise and ambiguity} in the datasets. We found that CRIC has a non-negligible amount of label noise (see Figs.~\ref{fig:noise-examples}, \ref{fig:noise-plot-cric}) across all the splits, which caused the models to learn erroneous patterns in the data. AOKVQA is comparatively less noisy but follows a similar noise distribution to CRIC (see Fig. \ref{fig:noise-plot-aokvqa}). This necessitates the need for external commonsense knowledge, while building noise robustness to specialise the models in the commonsense VQA task.

\subsection{End-to-End NLKI Performance}
\label{sec:performance-analysis}

Table \ref{tab:main_accuracy_table} shows that the full pipeline with \textbf{Type-5} explanations (with DC, RC, TC, O, RFs as the parameters of the prompt) is the strongest variant. This echoes the scores in Table \ref{tab:explanation_quality_scores}, as \textbf{Type-5} rationales most closely match groundtruth explanation in both $n$-gram overlap and the semantic space than \textbf{Types 1–4}.
\begin{table}[!htpb]
\centering
\resizebox{\columnwidth}{!}{
\begin{tabular}{@{}ccccc@{}}
\toprule
Architecture & \begin{tabular}[c]{@{}c@{}}Retrieved/\\ Generated\end{tabular} & \multicolumn{3}{c}{Accuracy} \\ 
\cmidrule(lr){3-5} 
 & & e-SNLI-VE & CRIC & AOKVQA \\ 
\midrule
KAT & Expl. Knowledge & 73.22 & 67.38 & 32.72\\
KAT & Expl. + Impl. Knowledge & \textbf{76.30} & \textbf{69.71} & \textbf{38.60}\\
\midrule
ViLT & \ding{55} & 76.46 & 72.99 & 24.01\\
Concat (ViLT) & \textbf{Type 5} & \cellcolor{gray!20}78.46 & \cellcolor{gray!20}74.95 & \cellcolor{gray!20}28.15\\
Concat (ViLT) & Type 7 & 64.66 & 63.34 & 20.08\\
Majority Voting (ViLT) & \textit{Q/CB-FT} & 74.51 & 69.08 & 13.45\\
\arrayrulecolor{gray!50}\midrule 
Concat (ViLT)+*CE & \textbf{Type 5} & \textbf{78.57} & \textbf{76.98} & \textbf{33.45} \\
\arrayrulecolor{black}\midrule
VisualBERT & \ding{55} & 74.48 & 62.60 & 23.6\\
Concat (VisualBERT) & \textbf{Type 5} & \cellcolor{gray!20}78.83 & \cellcolor{gray!20}64.69 & \cellcolor{gray!20}35.40  \\ 
Concat (VisualBERT) & Type 7 & 62.46 & 55.22 & 20.00 \\ 
Majority Voting (VisualBERT) & \textit{Q/CB-FT} & 72.53 & 60.25 & 19.00 \\
\arrayrulecolor{gray!50}\midrule 
Concat(VB)+*CE & \textbf{Type 5} &\textbf{78.95} & \textbf{67.15} & \textbf{40.12} \\
\arrayrulecolor{black}\midrule
FLAVA & \ding{55} & 79.93 & 73.11 & 33.07 \\
Concat (FLAVA) & \textbf{Type 5} & \cellcolor{gray!20}81.54 & \cellcolor{gray!20}75.02 & \cellcolor{gray!20}47.85\\ 
Concat (FLAVA) & Type 7 & 70.18 & 64.30 & 32.09 \\ 
Majority Voting (FLAVA) & \textit{Q/CB-FT} & 72.53 & 60.25 & 32.68 \\
\arrayrulecolor{gray!50}\midrule 
Concat(FLAVA)+*CE & \textbf{Type 5} & \textbf{82.05} & \textbf{77.85} & \textbf{47.85} \\
\arrayrulecolor{black}\midrule
\multicolumn{5}{c}{\textbf{Gold-label baseline}} \\ 
\arrayrulecolor{black}\midrule
ViLT & Ground Truth Expl. & 92.47 &  82.98  & 56.31 \\
ViLT & Type-0 & \textbf{97.91} & \textbf{97.02} & \textbf{58.38} \\
ViLT & Type 1 & 75.92 & 55.65 & 20.28 \\
\arrayrulecolor{black}\midrule
VisualBERT & Ground Truth Expl. & 93.49 & 77.00 & 69.35\\ 
VisualBERT & Type-0 & \textbf{97.89} & \textbf{96.60} & \textbf{87.16} \\ 
VisualBERT & Type 1 & 76.11 & 53.82 & 35.19\\
\arrayrulecolor{black}\midrule
FLAVA & Ground Truth Expl. & 93.04 &  82.36  & 72.65\\ 
FLAVA & Type-0 & \textbf{98.11} & \textbf{97.34} & \textbf{90.03} \\ 
FLAVA & Type 1 & 69.29 & 50.45 & 39.71 \\
\arrayrulecolor{black}\bottomrule
\end{tabular}
}
\caption{Performance of Baseline vs NLKI (ours) vs Gold-label baseline. Expl. and Impl. refer to \textit{Explicit} and \textit{Implicit} respectively. \textit{Q/CB-FT} are facts retrieved by fine-tuned ColBERTv2 with the question as the query, \textit{*CE} rows contain the best performance when various noise robust loss function is applied with the \textbf{Type-5} explanation integration (see \S\ref{sec:noise-robust}).}
\label{tab:main_accuracy_table}
\end{table}
The payoff is large: \emph{+13\%} for FLAVA on AOKVQA and \emph{+2\%} even on the noisy CRIC split. We also compared our \emph{NLKI pipeline} with a specialised retrieval-augmented baseline, termed KAT \cite{kat-framework}. The details of the knowledge concatenation and truncation strategy in our pipeline are present in Appendix \ref{sec:truncation}. We trained the Knowledge-Augmented Transformer (KAT) with explicit and implicit knowledge; KAT\footnote{Implementation details in Appendix \ref{sec:kat-implementation}.} lags behind the NLKI \textbf{Type-5} variant on every dataset except ViLT with \textbf{Type-5}. 

\paragraph{Effect of Noise.} Further analysis shows that NLKI still inherits a lot of label noise. We manually analysed $1000$ CRIC training samples (Fig. \ref{fig:noise-plot-cric}). This revealed five noise classes: label, image, question, image–question mismatch, and ambiguous labels (\S\ref{sec:noise-description})—with label noise being the most frequent ($180$ among $1000$). For example, the aircraft image in Fig. \ref{fig:noise-examples}c is labelled ``Grass'' instead of ``Water'', so even a correct \textbf{Type-5} explanation fails; Fig. \ref{fig:noise-examples}d allows two valid answers, blurring the target. Noise compounds when hallucinated explanations reinforce wrong labels (Fig.\ref{fig:hallucinated_explanation_collection}d). We also clearly see that the datasets have a varying degree of noise. On a similar analysis, nearly $90$ instances (out of $1000$) in AOKVQA answers were unanimously judged ambiguous or incorrect by three independent raters.\footnote{Details in Appendix~\ref{sec:noise-description}.} 
In contrast, e‐SNLI-VE natural language inference benchmark contains three NLI answer labels, considerably less noisy than VQA datasets. Hence, we next incorporate noise-robust loss functions to prevent small 240M VLMs from over-fitting to noisy outliers in \S\ref{sec:noise-robust}. With noise-robust loss functions, all architectures surpass KAT's performance by a large margin (except ViLT on AOKVQA), underscoring the competitiveness of our lightweight scheme.

\subsection{Building Noise Robustness}
\label{sec:noise-robust}
We replace the default cross‐entropy (CE) objective with \emph{down‐weight} or \emph{smoothen} the contribution of mislabeled samples. For each example image and question ($x_i$), target answer ($y_i$), we use the standard cross-entropy loss objective to fine-tune each model. Assume, for $i^{th}$ example, the model logits are $z_i\!\in\!\mathbb{R}^{C}$ ($C$=\#classes), $p_i=\mathrm{softmax}(z_i)$ and $p_{y}^{(i)}$ denote the predicted probability of the correct class. The standard loss becomes $\mathcal{L}_{\mathrm{CE}}\;=\;-\sum_i\log p_{y}^{(i)}.$ The noise-robust variants can be defined as follows.
\\\noindent
\textbf{Symmetric Cross‐Entropy (SCE).}~~
Following \citet{symmetriccrossentropy}, we combine the usual cross-entropy with a \emph{reverse} term that penalizes over-confident errors:
\begin{align}
\mathcal{L}_{\mathrm{RCE}}(p,y)
      &= -\sum_{c=1}^{C} p_{c}\,\log y_{c}
         \;=\; -A\!\!\sum_{c\neq y} p_{c} \\[2pt]
      &\quad =\; -A\,(1-p_{y}), \label{eq:rce}\\[4pt]
\mathcal{L}_{\mathrm{SCE}}(p,y)
      &= \alpha\,\mathcal{L}_{\mathrm{CE}}(p,y)
         + \beta\,\mathcal{L}_{\mathrm{RCE}}(p,y), \label{eq:sce}
\end{align}

where \(p_{y}\) is the predicted probability of the ground-truth class, we fix \(\log 0 = -\gamma\) with \(\gamma = 4\), \(\alpha = 0.1,\;\beta = 1.0\) (the values of $\alpha$ and $\beta$ are directly taken from \cite{symmetriccrossentropy}). The CE component ensures fast convergence, while the \(-\gamma(1-p_{y})\) term flattens the loss landscape around noisy labels, granting robustness to the \SIrange{10}{30}{\percent} label noise observed in CRIC and AOKVQA.
\\\noindent
\textbf{Generalised Cross‐Entropy (GCE).}~~
\citet{generalizedcrossentropy} proposed
\begin{equation*}
\mathcal{L}_{\mathrm{GCE}}(p,y)\;=\;\frac{1-p_{y}^{q}}{q},\quad 0<q\le1,
\label{eq:gce}
\end{equation*}
which reduces to mean absolute error (MAE) as $q\!\to\!0$ and to standard CE as $q\!\to\!1$.  
We use $q\!=\!0.7$ directly from \cite{generalizedcrossentropy}, and form a convex mixture with CE:
\begin{equation*}
\mathcal{L}_{\mathrm{Mixed}}(p,y)\;=\;\lambda\,\mathcal{L}_{\mathrm{CE}}(p,y)\;+\;(1-\lambda)\,\mathcal{L}_{\mathrm{GCE}}(p,y).
\label{eq:cegce}
\end{equation*}
We use $\lambda\!=\!0.4$ for noisy datasets (with $0.9$ for e-SNLI-VE, which has limited label noise).

\paragraph{Training protocol.}
For each loss function, we train with vanilla CE for two epochs to stabilise early gradients, then switch to the chosen robust loss. We keep the batch size, optimiser, and learning‐rate schedule identical to the CE baseline to isolate the effect of the loss.

\subsection{Effect of Noise-robust Training}

Table~\ref{tab:cric-loss-comparison} compares CE, SCE, and CE + GCE on all three datasets for ViLT, VisualBERT and FLAVA for CRIC.  
On \textbf{CRIC}, \textbf{SCE} lifts FLAVA by \textbf{+2.8\%}, VisualBERT by \textbf{+2.46\%} and ViLT by \textbf{+2.03\%} over CE, confirming that SCE’s reverse term curtails over-fitting to noisy labels. On \textbf{AOKVQA}, \textbf{SCE} delivers the best trade-off (\textbf{+5.5\%}), having low to moderate noise with ViLT and VisualBERT (Table \ref{tab:aokvqa-loss-comparison} in the appendix). But FLAVA already reaches high performance with vanilla CE, so the extra regularisation offered by SCE or GCE brings only a marginal change (-0.12 \%) and can even over-smooth (GCE+CE). On \textbf{e‐SNLI-VE} standard CE \emph{outperforms} its robust siblings. Raising $q$ (\(0.7\!\to\!0.95\)) or $\lambda$ (\(0.4\!\to\!0.9\)) makes CE + GCE converge to CE and closes the gap (Table~\ref{tab:esnlive-loss-comparison} in the appendix).

\begin{table}[!ht]
\centering
\resizebox{\columnwidth}{!}{%
\begin{tabular}{@{}lcccc@{}}
\toprule
Architecture & Type & CE & SCE & GCE+CE \\
\midrule
ViLT         & Type-5 & 74.95 & \textbf{76.98} & 75.13 \\
VisualBERT   & Type-5 & 64.69 & \textbf{67.15} & 65.66 \\
FLAVA        & Type-5 & 75.02 & \textbf{77.85} & 76.98 \\
\bottomrule
\end{tabular}%
}
\caption{Comparison of performance (in percentage) for different architectures using various loss functions on the CRIC test split of 76K samples. For e-SNLI-VE and AOKVQA, refer to Tables \ref{tab:esnlive-loss-comparison} and \ref{tab:aokvqa-loss-comparison} in the Appendix.}
\label{tab:cric-loss-comparison}
\end{table}
 
In our noise ablation (Figs. \ref{fig:noise-plot-cric} \& \ref{fig:noise-plot-aokvqa}, Appendix \ref{sec:noise-description}), SCE delivered the largest gains in conditions of high label noises, mirroring the threshold reported in \citet{symmetriccrossentropy}—whereas a 0.4 CE + 0.6 GCE mix performed the best in moderate level noise conditions, and standard CE remained optimal on e-SNLI-VE with only 3 labels. These findings explain why prior work that applied a single loss across heterogeneous benchmarks reported inconsistent gains; by adapting the loss to the dataset (according to the level of label-noise), we achieve stable improvements without any architectural change or extra inference cost.

\section{Performance of Generative VLMs}
\begin{table}[!ht]
\centering
\resizebox{\columnwidth}{!}{%
\begin{tabular}{@{}cccccc@{}}
\toprule
Models & EM & P & R & F1 & ACC \\ 
\midrule
Qwen2-VL & 31.18 & 40.00 & 92.82 & 93.17 & 41.90\\
SmolVLM & 10.48 & 31.42 & 89.91 & 90.54 & 33.89\\
MiniCPM & \textbf{58.60} & \textbf{96.05} & \textbf{95.7} & \textbf{95.83} & \textbf{58.58}\\
Phi3-Vision & 52.64 & 96.06 & 95.28 & 95.61 & 53.24\\ 
\bottomrule
\end{tabular}%
}
\caption{Performance of Generative Models on AOKVQA val split where \textit{EM} is exact string match \textit{P} is Precision, \textit{R} is Recall, \textit{F1} is the F1 score, and \textit{ACC} is the cosine similarity score.}
\label{tab:generative-models-aokvqa}
\end{table}

We further benchmark (smaller, less than $4B$) generative instruction-tuned models such as Qwen2-VL ($2B$) \cite{qwenVL}, Phi3-Vision ($4.1B$) \cite{phi-3-vision}, MiniCPM ($3.43B$) \cite{miniCPM} and SmolVLM ($2.25B$) on AOKVQA. From the results in Table~\ref{tab:generative-models-aokvqa}, we observe that \textbf{NLKI + SCE} turns a $240M$-param FLAVA into a model that beats 2B-param Qwen-VL and SmolVLM on AOKVQA with far less compute time and power.

\section{Discussion}
\paragraph{Learnings from Gold-label baseline Setup and Effect of Additional Context on Accuracy.} 
Table \ref{tab:main_accuracy_table} confirms that label-supervised \textbf{Type-0} explanations push accuracy far beyond both ground-truth texts and \textbf{Type-1} (no-label) variants. Injecting the answer often \textit{repairs} missing context—see Fig.\ref{fig:explanation_collection}b—and yields more assertive wording (e.g., “\emph{the macaroni is the main course}’’ in Fig.\ref{fig:explanation_collection}d). Yet the benefit is fragile: in a 1.5K sample drawn across all datasets, 51\% of \textbf{Type-0} explanations still hallucinate or add spurious detail because they rely on weak visual cues. By contrast, our gold-label-free \textbf{Type-5} explanation prompt, which feeds dense and region captions plus retrieved facts, cuts that rate to 18.5\% and grounds over 80\% of explanations in the scene (Fig. \ref{fig:hallucinated_explanation_collection}; definition in Appendix \ref{sec:florence-fn-description}). A Florence-finetuned captioner used in place of the LLM underperforms; its explanations are shorter, omit causal links, and have lower end accuracy (Table \ref{tab:explanation_quality_scores}; examples in Fig. ~\ref{fig:florence-ft-expl}).

\paragraph{When does knowledge integration help?}
Table \ref{tab:main_accuracy_table} demonstrates that \emph{NLKI alone without noise mitigation}, particularly with \textbf{Type-5} explanations, has a marginally positive impact on model performance for CRIC, while having a moderate to significant impact in e-SNLI-VE \& AOKVQA. Some key observations emerge from our analysis of the NLKI framework: \textbf{1.} \textit{The better the generated explanations are, the better the effectiveness of NLKI, which even outperforms the complex integration pipelines like KAT}. \textbf{2.} \textit{To frame a quality Natural Language knowledge that can be effectively utilised, the explanations should capture relevant visual context in textual form while minimising hallucinations}. \textbf{3.} \textit{NLKI, when coupled with noise robust losses, matches and even outperforms some of the performance of the architectures in the range of 1-4 B parameters}. \textbf{4.} \textit{The cleaner the dataset and the stronger the model is, the smaller the benefit of noise-robust losses due to less chance of overfitting on mislabeled samples}. For example, on AOKVQA, they mainly rescue the lighter architectures while leaving FLAVA nearly unchanged. \textbf{5.} \textit{If ground-truth (often noisy) labels are used to generate an explanation (gold-label baseline), they are more prone to hallucination, leading to unreliable predictions, as in Fig. \ref{fig:hallucinated_explanation_collection}}. If raw knowledge retrieved from knowledge bases or graphs is used as guidance, it may introduce additional noise and hinder performance rather than improving it. Table~\ref{tab:k-facts-concat} shows our ablation study on augmenting $k$ facts to questions degrades performance as $k$ increases.

\paragraph{Impact in the context of Generative VLMs.}
Table ~\ref{tab:generative-models-aokvqa} shows that despite rigorous pretraining and the use of advanced architectures, generative models still struggle with visual commonsense reasoning tasks. The best-performing model, MiniCPM, achieves only 58.60\%, followed by Phi-3 Vision. This highlights the persistent lack of commonsense knowledge in medium-sized VLMs, as they fail to outperform the NLKI coupled with the noise-robust loss framework when applied to encoder-only models such as FLAVA, ViLT, and VisualBERT. We plan to explore the impact of NLKI in generative models as part of future work.

\paragraph{Inference Time Latency of NLKI Framework.}
While we highlight NLKI’s empirical gains, it is also important to consider its computational footprint. Our framework introduces additional modules beyond the base VLM, namely, captioners, an object detector, a retriever and an explainer—so we report the full breakdown of latency, FLOPs, and GPU usage in the table below. The total pipeline latency for a single image–question pair is 1.32s when components are run sequentially, with most of the cost concentrated in captioning and explanation generation. The reader stage itself is negligible ($\leq 65$ ms across all tested VLMs). Running captioning and object detection concurrently shortens the critical path to 0.87s ($\approx 34\%$ faster), though this raises peak memory load from $\approx 5$ GB to $\approx 15$ GB. Importantly, the retriever and captioners can be executed offline or on CPU, and the explainer can be swapped for a smaller LLM (e.g., Llama 1B or 3B), further reducing the online GPU footprint. Thus, although NLKI involves multiple stages, its design remains lightweight and deployable, offering a favourable trade-off between efficiency and the substantial performance gains reported in \ref{results}.

\begin{table}[!ht]
\centering
\small
\setlength{\tabcolsep}{3pt} 
\renewcommand{\arraystretch}{1.1} 
\begin{tabular}{lccc}
\hline
\textbf{Tasks} & \textbf{Latency} & \textbf{FLOPS} & \textbf{GPU Usage} \\
\hline
Dense Cap. & 235.80 & 1680 & 5.2\\
Region Cap. & 313.75 & 1680 & 5.2\\
Object Det. & 225.06 & 1680 & 5.2\\
Expl. Gen. & 486.53 & 735.48 & 15\\
KB Retrieval & 114.12 & 7 & 0.78\\
Answer Prediction & 65.02 & 55.84 & 1.9\\
\hline
\multicolumn{4}{l}{\textbf{Overall Pipeline Latency:}} \\
\multicolumn{4}{l}{(235.80 + 313.75 + 225.06 + 486.53 + 65.02) = 1.32 sec} \\
\hline
\end{tabular}
\caption{GPU usage (in GB), wall-clock latency (in ms), and theoretical FLOPs (in GFLOPS) for the NLKI framework. \textit{Cap.} stands for Captioning, \textit{Det.} stands for Detection, \textit{Expl. Gen.} stands for Explanation Generation.}
\label{tab:latency}
\end{table}

\section{Acknowledgements}
We gratefully acknowledge the full support of the Science and Engineering Research Board (SERB), Government of India, through the Start-up Research Grant SRG/2022/000648 (PI: Somak Aditya) for this work.

\section{Conclusion}
Our study shows that pairing external commonsense knowledge with noise-aware training lifts $\leq$240M vision-language models to the tier of far larger systems. A lightweight NLKI pipeline-- ColBERT-v2 retrieval plus Llama-3.1-8B generated \textbf{Type-5} explanations-- boosts the performance of ViLT from 24 → 38 and FLAVA from 46 → 54 \% on AOKVQA, outmatching 1–4B parameter generative baselines at a fraction of the compute. Noise audits reveal 17–23\% faulty labels in CRIC and AOKVQA (virtually none in e-SNLI-VE); swapping vanilla CE loss for Symmetric CE adds up to +3\%, while a CE + GCE mix is best at moderate noise, and vanilla CE suffices for clean data. Finetuned ColBERT raises retrieval precision, and enriching prompts with dense/region captions curbs hallucination, though one fact per question is optimal given VLM context limits. In sum, NLKI plus a dataset-aware robust loss turns 250M-parameter sVLMs into noise-resilient commonsense VQA engines that rival or beat multi-billion-parameter models without extra inference cost.

\section*{Limitations}
Our pipeline is modular, but still serial retriever, explainer and reader are tuned in isolation; we do not explore joint optimisation. Noise robustness is tackled only at the label level with loss re-weighting; we leave open richer defences against other noises (such as in questions, image-question mismatch, etc). Our explanation module relies on Llama-3.1-8B, a single LLM family; whether larger or instruction-specialised models change the trade-offs is unknown. NLKI assumes access to accurate object lists and dense/region captions from auxiliary detectors, but the impact of errors in these vision tools is not assessed directly.

\bibliography{main}
\newpage
\section*{Appendix}
\appendix
\section{Datasets}
\label{sec:dataset-descriptions}
We focus on CRIC, AOKVQA, and e-SNLI-VE, which emphasise open-ended, contextual commonsense reasoning over encyclopedic or synthetic factoid knowledge. Datasets like FVQA and WHOOPS, which centre on factual recall or synthetic errors, are excluded as they diverge from our goal of modelling everyday reasoning.

\textbf{CRIC} contains 494K automatically generated questions over 96K Visual Genome images, enriched with 3.4K knowledge items from ConceptNet and Wikipedia.
\textbf{AOKVQA} includes ~25K diverse, crowdsourced questions based on COCO images. Both CRIC and AOKVQA are multiple-choice tests with four options each.
\textbf{e-SNLI-VE} combines Flickr30k and SNLI-VE, with hypotheses labeled as \textit{Entailment}, \textit{Contradiction}, or \textit{Neutral}.

\begin{table}[!ht]
\centering
\resizebox{\columnwidth}{!}{%
\begin{tabular}{@{}cccccc@{}}
\toprule
Dataset & Type & Train & Val & Test & Answer Type \\ 
\midrule
CRIC & VQA & 364K & 76K & 84K & MCQ\\
AOKVQA & VQA & 17K & 1.1K & 6.7K & MCQ\\
e-SNLIVE & NLI & 401K & 14K & 14K & NLI Labels\\
\bottomrule
\end{tabular}%
}
\caption{Table showing the split size and answer type for all datasets.}
\label{tab:datasets-used}
\end{table}

\section{Architecture Description}
\label{sec:models-used}
We have selected models all under 240M for our VQA classification task. We opted for two different architectures 1. Single Stream 2. Dual Stream is based on the processing of both modalities, vision and text.
\\\noindent
\textbf{ViLT} \cite{vilt1} is a Transformer-based unified architecture that uses a single transformer module to encode and fuse both visual and text features in place of a separate deep visual embedder. 
\\\noindent
\textbf{VisualBERT} \cite{li2019visualbert} requires a set of visual embeddings (representing the visual features)  extracted from an image encoder and text embeddings encoded with BERT. Then, it utilises the self-attention mechanism for implicit alignment of elements in the input text and regions within the input image. 
\\\noindent
\textbf{FLAVA} \cite{flava} is a multimodal transformer-based model designed to process and align visual and textual information for various vision and language tasks.

\paragraph{KAT Implementation}
\label{sec:kat-implementation}
KAT originally used filtered English Wikidata and GPT-3 to extract implicit knowledge. In our setup, we use \verb|gpt-3.5-turbo| for this step. Since VinVL and Oscar (used by KAT for extracting image features and captions) are no longer publicly available, we replace them with Florence-generated dense and region captions to query GPT-3.5. We retain the original prompt format but, due to resource constraints, combine answer and explanation generation into a single API call with a modified prompt. Final results are reported in Table~\ref{tab:main_accuracy_table}.

\paragraph{Model Context Length \& Truncation Strategy}
\label{sec:truncation}
To integrate external knowledge into the model input, we prepend the generated explanation to the original question using the format: \verb|<explanation>[SEP]<question>| before pairing this text with the corresponding image features. For the smaller reader VLMs, we enforce a strict 100-token limit on the concatenated explanation–question string: if the combined text exceeds this budget, truncation is applied from the end, ensuring that the leading tokens, which typically contain the most informative content, are preserved. This strategy was consistently applied across all datasets and models. Empirically, truncation was seldom triggered, as the average token counts remained well below the limit (e.g., CRIC: 12.74 tokens for questions and 19.35 for explanations; AOKVQA: 8.69 and 9.09; e-SNLI-VE: 7.39 and 10.43, respectively).

\section{Retrieval}
\paragraph{What is majority voting?}
\label{majority}
In the majority voting setup, we concatenate each of the top-5 retrieved facts individually with the question and image context to form five separate inputs. The model then predicts an answer for each of these fact-augmented inputs independently. We collect all five predicted labels and select the final answer by majority vote—i.e., the label that appears most frequently across the five runs. This approach helps smooth over noisy or irrelevant facts by relying on the consensus across multiple evidence sources, making the final prediction more robust to individual retrieval errors.
\paragraph{Comparison of Retrieval Performance from OMCS 1.5M and 20M Knowledge Corpus} 
To verify the commonsense facts quality, we chose a random sample of 20K from both the CRIC and e-SNLI-VE datasets. Although the CRIC test set is larger, a subset of 20K facts was chosen to make the results comparable to those of e-SNLI-VE, which has 15K samples in the test set. Looking at the results, it is evident that the OMCS corpus alone contains more relevant facts for commonsense reasoning when compared to the Ground Truth explanations. OMCS is also nearly 20 times smaller; hence, retrievals from the corpus are significantly faster.
\begin{table*}[ht]
\small
\centering
\begin{tabular}{@{}ccccccc@{}}
\toprule
Corpus & K & Rouge1 & Rouge2 & RougeL & BLEU & Cosine Score \\ 
\midrule
 & @ 5 & 55.86 & 45.52 & 54.94 & 57.68 & 60.58 \\
\multirow{-2}{*}{OMCS 1.5M} & @ 10 & 58.86 & 49.22 & 58.09 & 66.74 & 62.38 \\
\midrule
 & @ 5 & 41.06 & 28.11 & 39.69 & 51.49 & 45.63 \\
\multirow{-2}{*}{20M Comm. Corpus} & @ 10 & 38.96 & 25.48 & 37.56 & 61.19 & 43.74 \\ 
\midrule 
\end{tabular}
\caption{BLEU, ROUGE, and Cosine Score of facts retrieved from the OMCS Corpus against facts retrieved from the cumulative 20M corpus introduced by \cite{raco} using pre-trained ColBERTv2. Retrieved facts are compared to the ground truth explanation. Scores were reported on 20K random samples from the test set of e-SNLI-VE.}
\label{tab:cric_omcs_vs_20M_corpus}
\end{table*}

\begin{table*}[ht]
\small
\centering
\begin{tabular}{@{}ccccccc@{}}
\toprule
Corpus & K & Rouge1 & Rouge2 & RougeL & BLEU & Cosine Score \\ \midrule
\multirow{2}{*}{OMCS 1.5M} & @ 5 & 28.58 & 13.71 & 25.86 & 48.12 & 30.67 \\
 & @ 10 & 28.52 & 13.71 & 25.77 & 47.96 & 15.37 \\
\midrule
\multirow{2}{*}{20M Comm. Corpus} & @ 5 & 27.84 & 12.50 & 24.81 & 41.63 & 26.21 \\
 & @ 10 & 27.63 & 12.49 & 24.95 & 41.84 & 13.17 \\ 
\midrule
\end{tabular}
\caption{BLEU, ROUGE, and Cosine Score of facts retrieved from the OMCS Corpus against facts retrieved from the cumulative 20M corpus introduced by \cite{raco} using pre-trained ColBERTv2. Retrieved facts are compared to the ground truth explanation. Scores were reported on 20K random samples from the test set of CRIC.}
\label{tab:esnli_omcs_vs_20M_corpus}
\end{table*}

\paragraph{Retrieval Performance for Commonsense VQA Tasks}
Here we present the detailed scores across the metrics like BLEU, ROUGE, and cosine similarity for top-5 and top-10 results for measuring the retrieval performance specific to our task. In Table. \ref{tab:retrieval_scores} we present the result of the three top retrievers for clarity, which was found to be effective for our commonsense VQA task.

\begin{table*}[!ht]
\small
\centering
\begin{tabular}{@{}ccccccc@{}}
\toprule
Query & Method & R-L@5 & R-L@10 & B-1@5 & Cosine@5 & Cosine@10 \\ 
\midrule
Q & SBERT + FAISS & \textbf{41.86} & \textbf{47.80} & \textbf{38.30} & \textbf{53.56} & \textbf{58.46} \\
C + Q & SBERT + FAISS & 38.57 & 44.70 & 35.21 & 51.54 & 56.88 \\
Objects + Q & SBERT + FAISS & 40.93 & 46.73 & 37.34 & 52.61 & 57.67 \\
SG + Q & SBERT + FAISS & 36.05 & 41.91 & 32.88 & 47.01 & 52.21 \\
\textit{All} + Q & SBERT + FAISS & 35.52 & 46.55 & 32.65 & 46.55 & 51.78 \\
\midrule
Q & ColBERTv2 & 61.33 & 67.32 & 56.24 & 68.69 & 73.11 \\
\midrule
Q & ColBERTv2-FT & \textbf{74.48} & \textbf{77.44} & \textbf{69.99} & \textbf{77.65} & \textbf{80.62} \\
C + Q & ColBERTv2-FT & 52.03 & 55.77 & 45.95 & 64.34 & 68.39 \\
Objects + Q & ColBERTv2-FT & 52.11 & 56.26 & 46.27 & 66.01 & 69.41 \\
SG + Q & ColBERTv2-FT & 31.12 & 34.89 & 27.56 & 41.99 & 43.75 \\
\textit{All} + Q & ColBERTv2-FT & 19.40 & 23.39 & 16.94 & 33.89 & 37.67 \\
\midrule
Q & Stella-en-v5 & \textbf{46.72} & \textbf{54.27} & \textbf{42.01} & 62.58 & 67.18 \\
C + Q & Stella-en-v5 & 32.28 & 38.76 & 28.56 & 50.14 & 55.10 \\
Objects + Q & Stella-en-v5 & 43.89 & 51.36 & 39.27 & \textbf{62.98} & \textbf{67.66} \\
SG + Q & Stella-en-v5 & 30.02 & 36.50 & 27.50 & 50.02 & 53.65 \\
\textit{All} + Q & Stella-en-v5 & 29.01 & 34.65 & 25.67 & 46.76 & 51.15 \\
\bottomrule
\end{tabular}
\caption{Comparison of BLEU, ROUGE, and Cosine Scores for Various Retrieval Mechanisms (on CRIC test split). We report the performance of different retrieval methods based on BLEU-1 (B-1), ROUGE-L (R-L), and Cosine similarity scores across different query settings. Finetuned ColBERTv2 outperforms all other methods. $Q$: Question, $C$: Caption, $O$: Detected Objects}
\label{tab:retrieval_scores}
\end{table*}
\paragraph{Examples of Retrieved Facts: FAISS vs ColBERTv2}
\begin{figure*}[!htb]
\centering
\includegraphics[width=\textwidth]{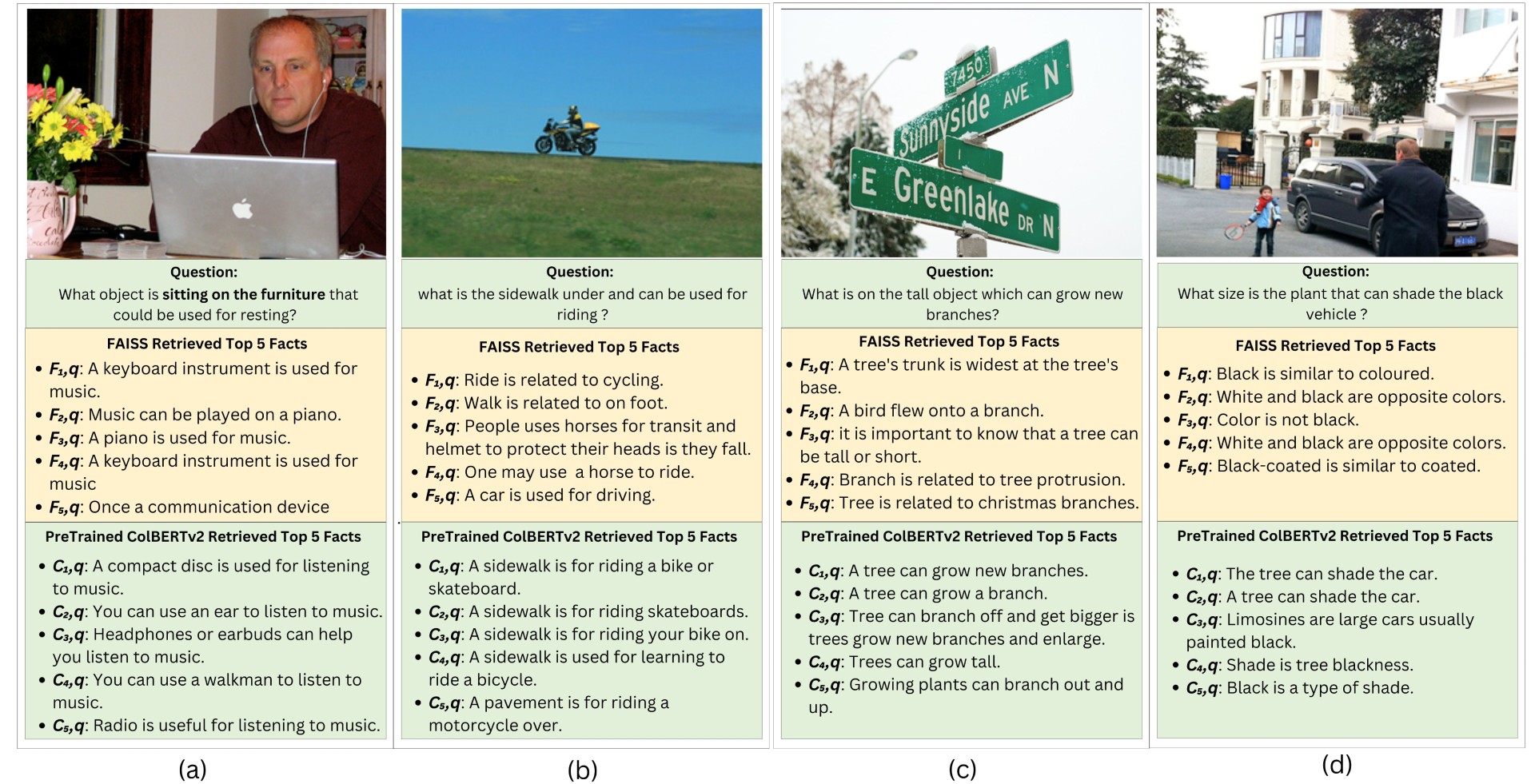} 
\caption{Examples illustrating that pre-trained COLBERTv2 performs better than FAISS vector search in retrieving contextually relevant facts from OMCS Corpus. $C_{i,q}$ indicates $i^{\text{th}}$ fact retrieved using the question as the query by the pre-trained ColBERTv2, $F_{i,q}$ indicates $i^{\text{th}}$ fact retrieved using a question as the query by FAISS vector search.}
\label{fig:pt_vs_ft_collection}
\end{figure*}
We provide more examples in Fig. \ref{fig:pt_vs_ft_collection}, depicting the difference in relevance of the facts retrieved by FAISS and those retrieved by Colbert. In (a), both FAISS and ColBERT retrieve facts unrelated to the query, which asks about \textit{furniture}, whereas the facts are about objects related to producing or listening to music. In (b), only the first fact retrieved by FAISS is close to what the query asks for, and all other facts are only connected to the query by one concept (such as `riding' or `walk'). The facts retrieved by ColBERT are directly related to sidewalks and activities it can be used for. Although none are direct answers, they are still semantically much closer to the query than the FAISS-retrieved facts. A similar pattern is noticed in (c), wherein the FAISS-retrieved facts are only related to the query by a singular concept (such as ``branch" or ``tree"), but the Colbert facts are both conceptually and contextually relevant; in this case, they are direct answers to the query. 

\paragraph{Finetuning Colbert}
\label{sec:finetuning-colbert}
Using a contrastive learning approach, we optimise the pairwise softmax cross-entropy loss over the relevance score $S_{q,d}$  for \( d^+ \) and \( d^- \) with the query $q$. 
\begin{equation*}
L(q, d^+, d^-) = -\log \left( \frac{\exp(S_{q,d^+})}{\exp(S_{q,d^+}) + \exp(S_{q,d^-})} \right)
\end{equation*}
We used a batch size of $32$, with a single $46$ GB NVIDIA A$40$ GPU, having a maximum length of the facts as $128$ (most facts are much smaller in length), and default settings for other parameters as in \citet{colbert}. We used $160K$ samples from the training splits of CRICs for fine-tuning due to the variety of queries in it. We experimented with adding more than $160K$ data to finetune it, but it didn't help improve the scores any further.

\section{Effect of Increasing Number of Retrieved Facts}
\begin{table}[!ht]   
\resizebox{\columnwidth}{!}{%
\begin{tabular}{@{}ccccc@{}}
\toprule
Architecture & Input & e-SNLI-VE & CRIC & AOKVQA \\ 
\midrule
\multirow{5}{*}{ViLT} & Question & \textbf{76.46} & \textbf{72.99} & \textbf{24.01}\\
 & 1 Fact + Question & 75.60 & 71.38 & 13.45 \\
 & 2 Facts + Question & 69.88 & 71.64 & 13.45 \\
 & 3 Facts + Question & 65.44 & 62.00 & 11.38 \\
 & 5 Facts + Question & 64.62 & 38.30 & 13.25 \\
\midrule
\multirow{5}{*}{VisualBERT} & Question & \textbf{74.48} & \textbf{62.60} & \textbf{23.6} \\
 & 1 Fact + Question & 74.34 & 61.00 & 22.0 \\
 & 2 Facts + Question & 73.99 & 20.25 & 20.0\\
 & 3 Facts + Question & 73.98 & 19.00 & 19.0\\
 & 5 Facts + Question & 73.62 & 19.00 & 19.0\\
\midrule
\multirow{5}{*}{FLAVA} & Question & \textbf{79.93} & \textbf{73.11} & \textbf{33.07}\\
 & 1 Fact + Question & 79.59 & 71.36 & 32.68 \\
 & 2 Facts + Question & 79.76 & 31.51 & 31.51\\
 & 3 Facts + Question & 79.81 & 31.31 & 34.48\\
 & 5 Facts + Question & 78.72 & 25.84 & 25.84\\
\bottomrule
\end{tabular}
}
\caption{Performance on varying the number of facts retrieved by finetuned ColBERTv2 concatenated with the query.}
\label{tab:k-facts-concat}
\end{table}
While somewhat useful, retrieved facts generally perform the worst in enhancing model performance, especially as more facts are concatenated. The primary hindrance is their often disjoint nature and lack of direct relevance to the specific question. In our experiments, as the number of concatenated facts increased, we consistently observed a decline in accuracy across various datasets. This decline can be attributed to several factors: the relevance of retrieved facts diminishes with quantity, introducing noise and reducing clarity. For instance, accuracy in the CRIC dataset dropped from 72.99\% with just the question to 38.30\% with five facts appended to the question. Similarly, ViLT's performance on the e-SNLI-VE dataset fell from 76.46\% to 69.62\%. The same pattern repeats for other models as well. The inclusion of more facts tends to clutter the input with irrelevant or redundant information, making it difficult for the model to focus on the relevant information, leading to degraded performance.

\section{Generative Small VLMs}
\label{sec:generative-vlm-description}
As the research community increasingly prioritises efficiency and low carbon footprints, there has been a growing shift towards small-vision language models (sVLMs) and low-resource training strategies. Therefore, in the context of commonsense VQA, we benchmarked the mentioned generative instruction-tuned VLMs\footnote{We used the standard HuggingFace implementations for all the small generative VLMs: \href{https://huggingface.co/Qwen/Qwen2-VL-2B-Instruct}{\url{https://huggingface.co/Qwen/Qwen2-VL-2B-Instruct}}, \href{https://huggingface.co/openbmb/MiniCPM-V}{\url{https://huggingface.co/openbmb/MiniCPM-V}}, \href{https://huggingface.co/HuggingFaceTB/SmolVLM-Instruct}{\url{https://huggingface.co/HuggingFaceTB/SmolVLM-Instruct}}, \href{https://huggingface.co/microsoft/Phi-3-vision-128k-instruct}{\url{https://huggingface.co/microsoft/Phi-3-vision-128k-instruct}}
} on our datasets. We limit ourselves to models with $<$ 4 billion parameters, as shown in Figure \ref{fig:accuracy-vs-size}. 
The effectiveness of these rigorously pre-trained generative models in commonsense VQA remained underexplored.
The observations in Table ~\ref{tab:generative-models-aokvqa} suggest the reason for the sub-optimal \textit{EM} and \textit{Cosine Accuracy} scores is that generative models can provide expressive answers based on their pre-training, but fail when asked questions that require explicit commonsense reasoning abilities. Also, it raises concerns about their ability to generalise beyond their training data, as in our case, no external knowledge was provided during this benchmarking. We used bert-score \cite{bertscore} to evaluate the \textit{P}, \textit{R}, \textit{F1}, which uses the contextual embeddings for computing token similarity. We chose the cosine similarity threshold as \textit{0.71} based on manual analysis of results specific to our datasets.

\section{Qualitative Analysis of Generated, Retrieved and Manual Explanations.}
\begin{figure*}[!htb]
\centering
\includegraphics[width=\textwidth]{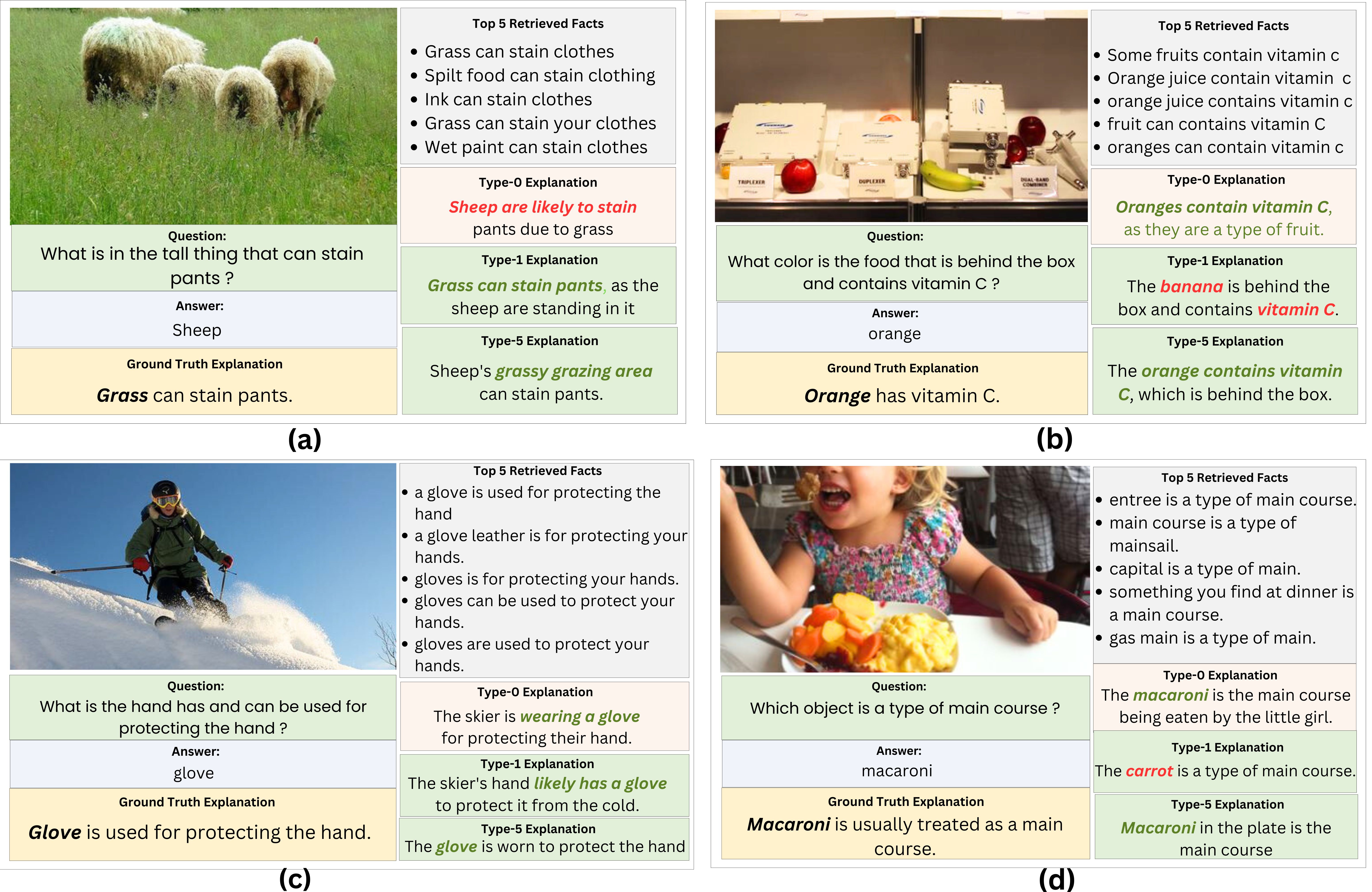} 
\caption{Examples from the CRIC dataset showing the difference between the Ground Truth, Type-0, Type-1 and Type-5 explanations generated by the LLama-3.1-8B. In all four cases, the \textbf{Type-5} explanation is the most rational and contextually relevant. \textbf{Type-0} and \textbf{Type-1} explanations are often hallucinated (to align the explanation with the label) or do not make rational sense. More examples are in the Appendix.}
\label{fig:explanation_collection}
\end{figure*}

In Fig.~\ref{fig:explanation_collection}, we show ground truth, retrieved facts, and generated explanations (\textbf{Types 0, 1, and 5}) comparatively for images in CRIC. Ground-truth explanations are accurate but lack context grounded in the query, making them less helpful for complex queries. \textbf{Type-1} generated explanations depend on the often incomplete context provided by the caption. For example, for Fig. \ref{fig:explanation_collection}b, the caption is ``there are many different types of food on display on the case'' and the objects include only ``bananas'', as the orange is obscured from view by the box in front. The LLM hallucinates and claims that the banana is behind the box, whereas it is actually in front. But the \textbf{Type-5} explanation, which includes region information, correctly detects the orange behind the box and reasons that an \textit{orange} has a higher vitamin C content and is also behind the box. 
Retrieved facts often consist of fragmented information that lacks the necessary context or specificity for the exact question. These facts may score higher on metrics like BLEU or ROUGE due to shared phrases with ground-truth explanations, but they often fail to contribute significantly to the model’s reasoning process, as they do not contain the elaborate context required for accurate answer derivation.

\section{Noise}
\paragraph{Noise in Datasets}
\label{sec:noise-description}
\begin{figure}[t]
\centering
\includegraphics[width=\linewidth]{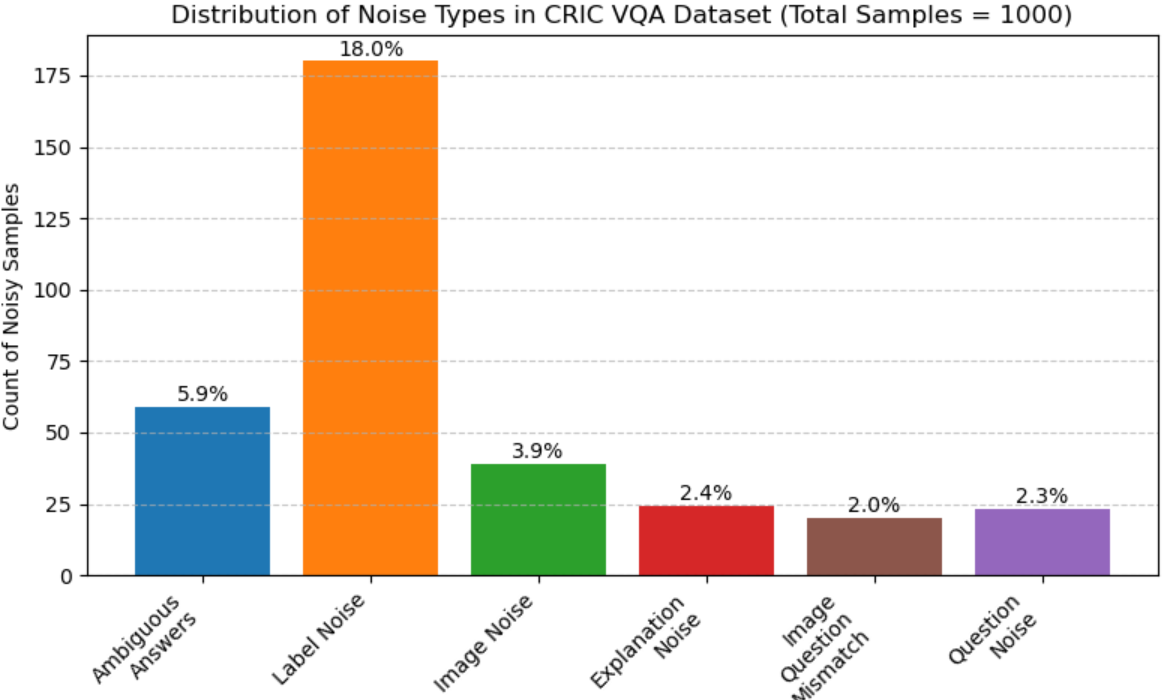} 
\caption{Noise Distribution of CRIC. We checked 1000 random samples, out of which 175 samples had label noise and others had the above distribution.}    
\label{fig:noise-plot-cric}
\end{figure}

\begin{figure}[t]
\centering
\includegraphics[width=\linewidth]{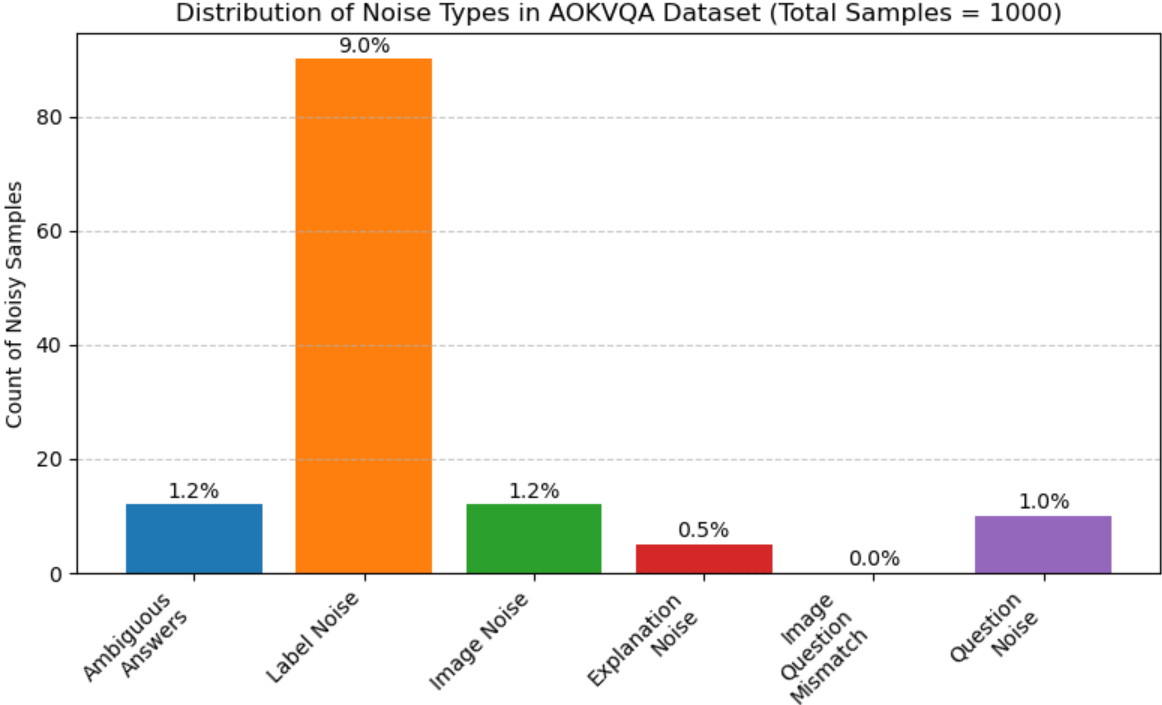} 
\caption{Noise Distribution of AOKVQA. We checked 1000 random samples, out of which around 90 samples had label noise, and others had the above distribution, which is significantly better than CRIC.}    
\label{fig:noise-plot-aokvqa}
\end{figure}

\begin{figure*}[!htb]
\centering
\includegraphics[width=\textwidth]{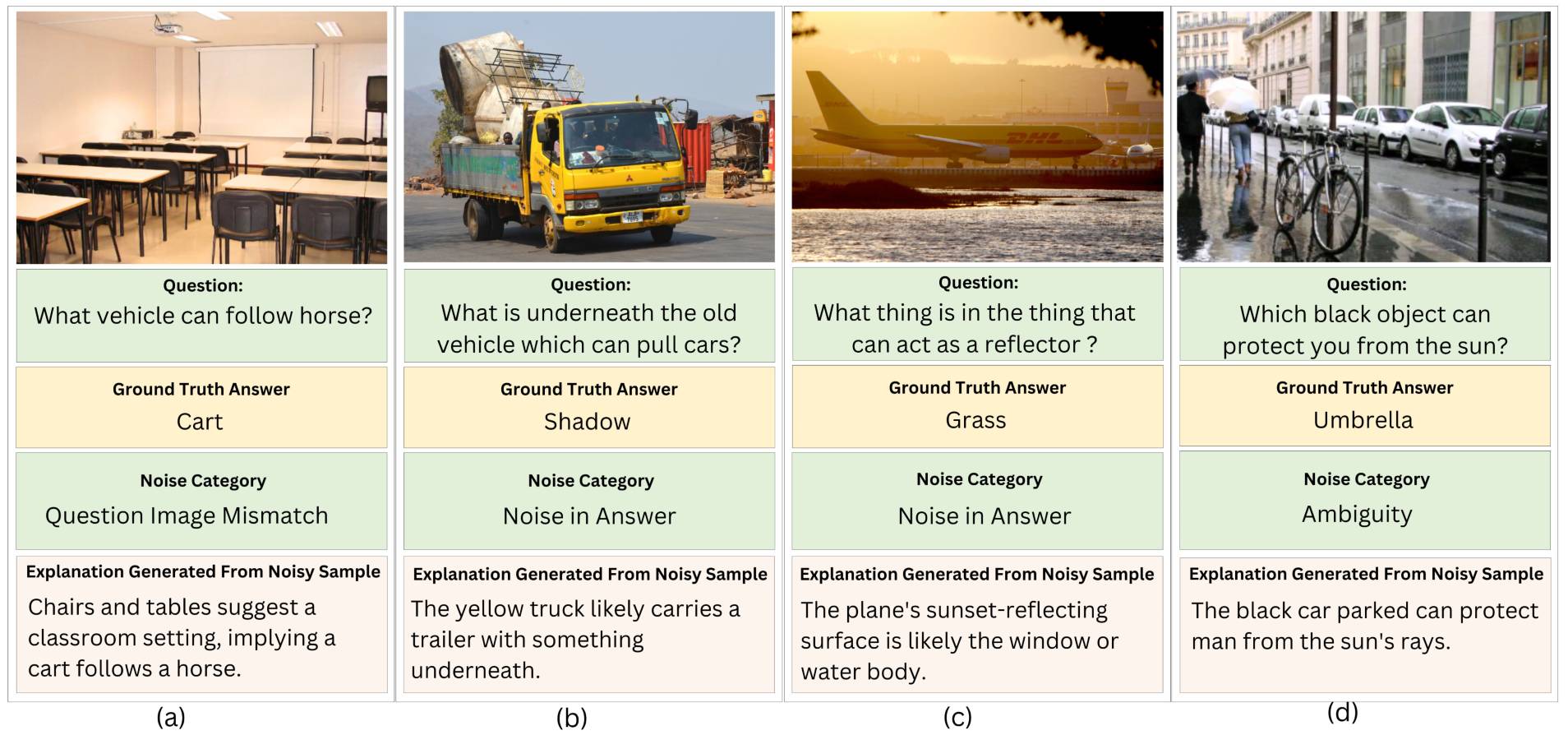} 
\caption{Examples illustrating the \textbf{noisy samples} from the datasets. where \textit{Noise Category} is the type of noise we encountered during manual analysis, \textit{Explanation Generated From Noisy Sample} is the LLaMA-3.1-8B explanation generated as a result of the noise for each sample, which degrades the overall performance. }
\label{fig:noise-examples}
\end{figure*}

Figure \ref{fig:noise-examples} presents noisy samples from our chosen dataset that contribute to performance degradation. Through manual analysis of a substantial portion of the dataset, we identified issues such as noisy labels, ambiguous answers, and unclear questions. Below, we highlight some noteworthy examples illustrating these noise types. Additionally, Figs. \ref{fig:noise-plot-cric} and \ref{fig:noise-plot-aokvqa} categorise the various types of noise present in the CRIC dataset.

\paragraph{Noise Ablations of e-SNLI-VE and AOKVQA}
Here we present the effect of using Symmetric Cross Entropy Loss and Generalised Cross Entropy Loss when applied to e-SNLI-VE and AOKVQA datasets in Tables~\ref{tab:esnlive-loss-comparison}\&\ref{tab:aokvqa-loss-comparison} respectively. Other than FLAVA, both SCE and GCE+CE combinations provide noticeable gains. For e-SNLI-VE, however, such gains are not prominent. Also, we present the noise distribution of CRIC and AOKVQA, analysed by 3 independent raters who are computer science graduates, two of whom are authors of this paper. 

\begin{table}[!ht]
\centering
\resizebox{\columnwidth}{!}{%
\begin{tabular}{@{}lcccc@{}}
\toprule
Architecture & Type & CE & SCE & GCE + CE \\
\midrule
ViLT         & Type-5 & 78.46 & 77.75 & \textbf{78.57} \\
VisualBERT   & Type-5 & 78.83 & 77.23 & \textbf{78.95} \\
FLAVA        & Type-5 & 81.54 & 80.47 & \textbf{82.05} \\
\bottomrule
\end{tabular}%
}
\caption{Comparison of performance (in percentages) for different architectures using various loss functions for e-SNLI-VE.}
\label{tab:esnlive-loss-comparison}
\end{table}

\begin{table}[!ht]
\centering
\resizebox{\columnwidth}{!}{%
\begin{tabular}{@{}lcccc@{}}
\toprule
Architecture & Type & CE & SCE & GCE + CE \\
\midrule
ViLT         & Type-5 & 28.15 & \textbf{33.45} & 32.33 \\
VisualBERT   & Type-5 & 35.40 & \textbf{40.12} & 38.51 \\
FLAVA        & Type-5 & \textbf{47.85} & 47.73 & 46.45 \\
\bottomrule
\end{tabular}%
}
\caption{Comparison of performance (in percentages) for different architectures using various loss functions for AOKVQA.}
\label{tab:aokvqa-loss-comparison}
\end{table}

\section{Explanation Generation}
\paragraph{Examples for Type-0, Type-1, Type-5 Explanations}

\begin{figure*}[!htb]
\centering
\includegraphics[width=\textwidth]{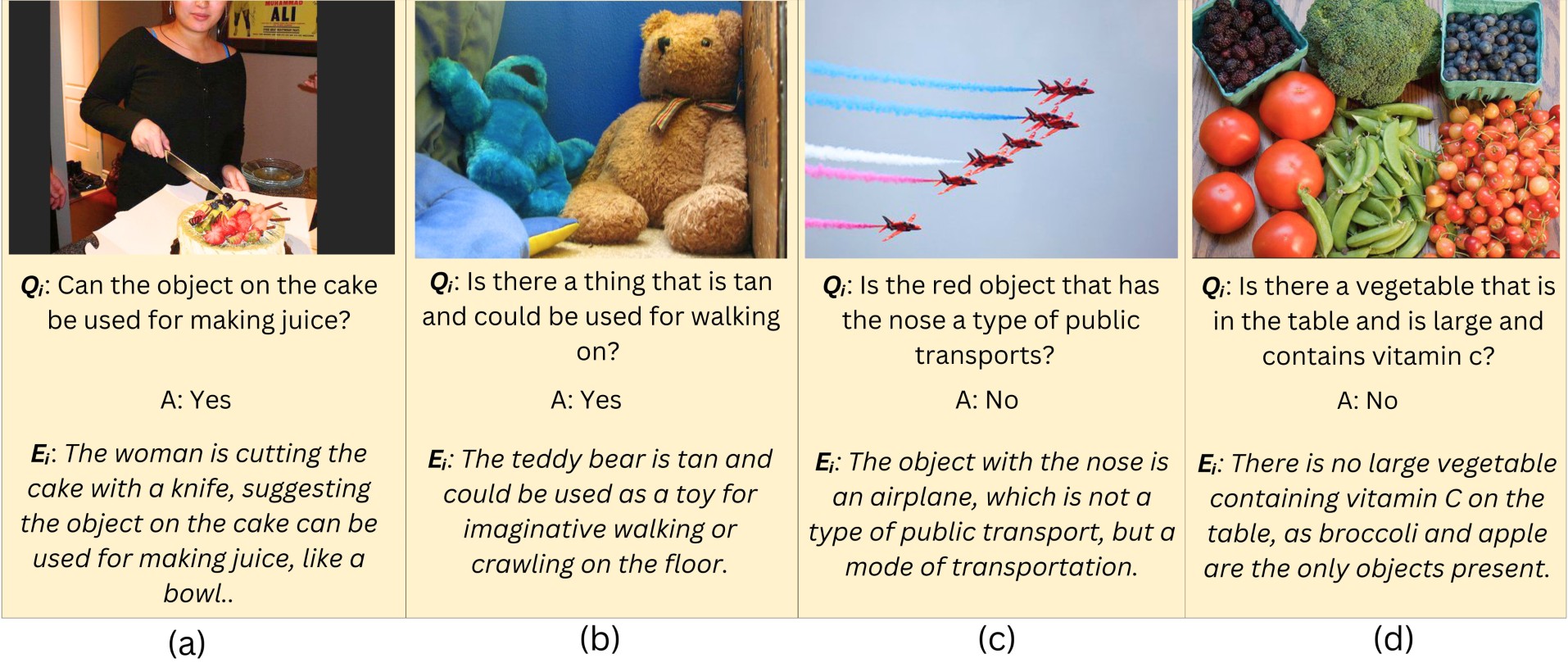} 
\caption{We report some instances where the LLM generates inaccurate \textbf{Type-0} explanations (where ground truth label was used). The visual and textual disconnects highlight the challenges LLMs face in reasoning, leading to 'hallucinations' that stray away from the visual context in the image.}
\label{fig:hallucinated_explanation_collection}
\end{figure*}

Here we provide more examples of the different \textbf{Types} of explanation produced by Llama-3.1-8B index (a - g). Of the 8 examples presented, only in (c) is the \textbf{Type-5} explanation incorrect. Even in that case, the explanation is still rationally correct because there is indeed no  ``object" on the aeroplane-- the question refers to the text ``FOX" on the engine, which is technically not an object. In all the examples, the \textbf{Type-0} explanations are not semantically consistent; they stray off the context very quickly and produce factually incorrect claims. This is evidenced by the examples in Fig.\ref{fig:hallucinated_explanation_collection}. Although consistent, the \textbf{Type-1} explanations are nearly never precise enough to enable the downstream VLM to identify the correct answer. 

\begin{figure*}[h!]
    \centering
    \includegraphics[width=\textwidth]{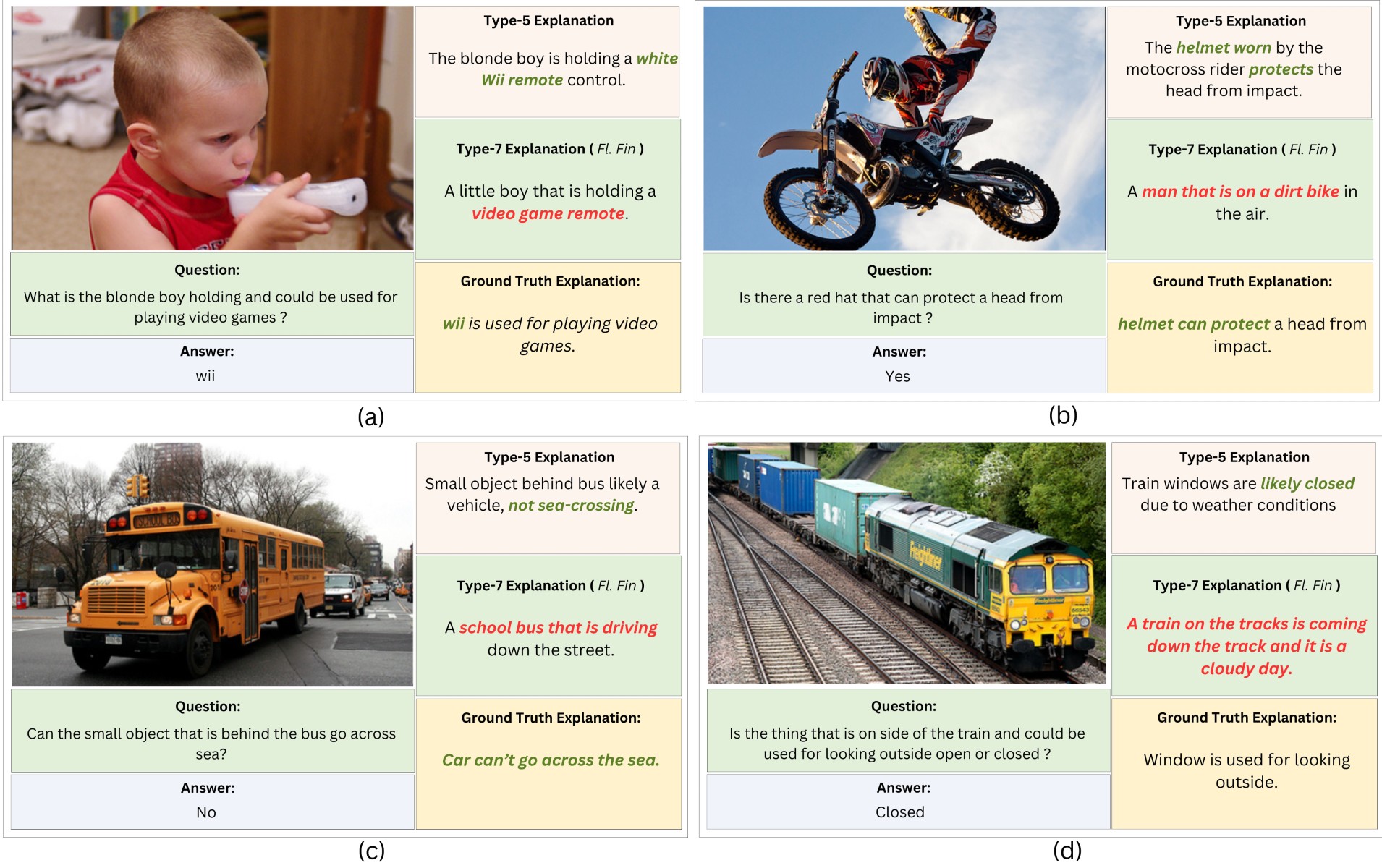} 
    \caption{Examples showing difference between the LLAMA-3.1-8B generated \textbf{Type 5} explanation and Florence Finetuned Generated explanations \textbf{Type 7}}
    \label{fig:florence-ft-expl}  
\end{figure*}

\paragraph{Florence Finetuning For Explanation Generation And Comparison.}
\label{sec:florence-fn-description}
As we explore the generation capability of Florence (sVLM) with 606M parameters, we found that, though it excels in several vision-language tasks, it failed to generate a well-reasoned explanation concerning the context of its image and the questions supplied to it. To ensure the quality of data to finetune Florence, we used a training split of the VQAe dataset comprising 181K data instances. We used VQAe \cite{vqae} because it closely resembles the type of explanations that CRIC and e-SNLI-VE have. We report the Florence finetuned performance on CRIC in \ref{tab:explanation_quality_scores}. In \ref{fig:florence-ft-expl} we show the difference between explanations generated by the Florence vs LLAMA-3.1-8B \textbf{(Type 5)}.

\paragraph{LLM Prompts for Explanation Generation: LLaMA-3.1-8B \& GPT-3.5}
We feed GPT-3.5 and LLaMA-3.1-8B with Image captions, Objects Detected, and Retrieved Facts from OMCS, then ask it to analyse and generate a short explanation leveraging these details. We kept on improving our prompts through experimentation because GPT/LLAMA generated extra texts containing additional reasoning for the explanation it generated, which are difficult to post-process. Below, we illustrate the prompt template.

\begin{figure*}[h!]
    \centering
    \includegraphics[width=\textwidth]{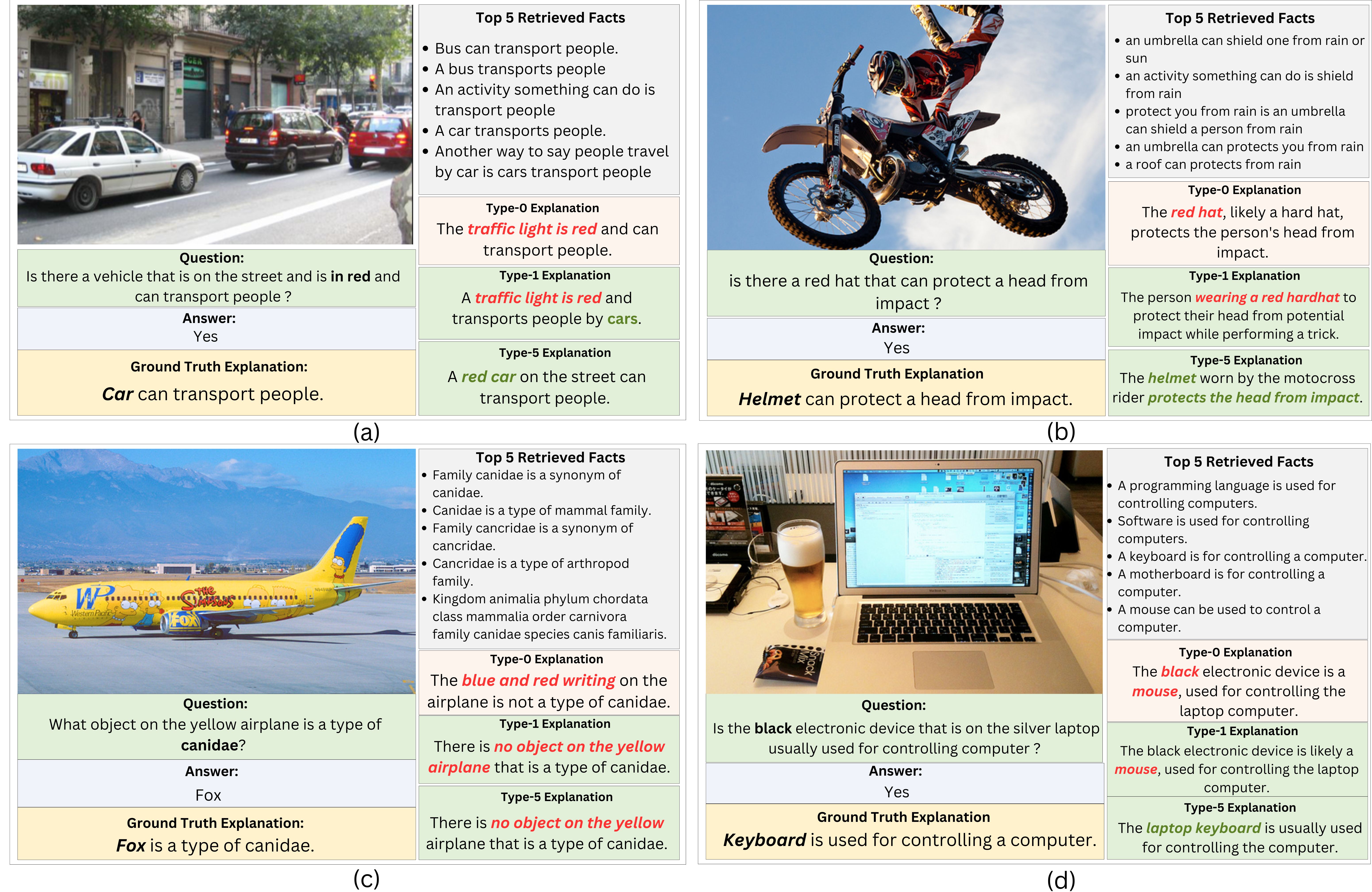} 
    \label{fig:explanation_collection2}    
\end{figure*}
\begin{figure*}[h!]
    \centering
    \includegraphics[width=\textwidth]{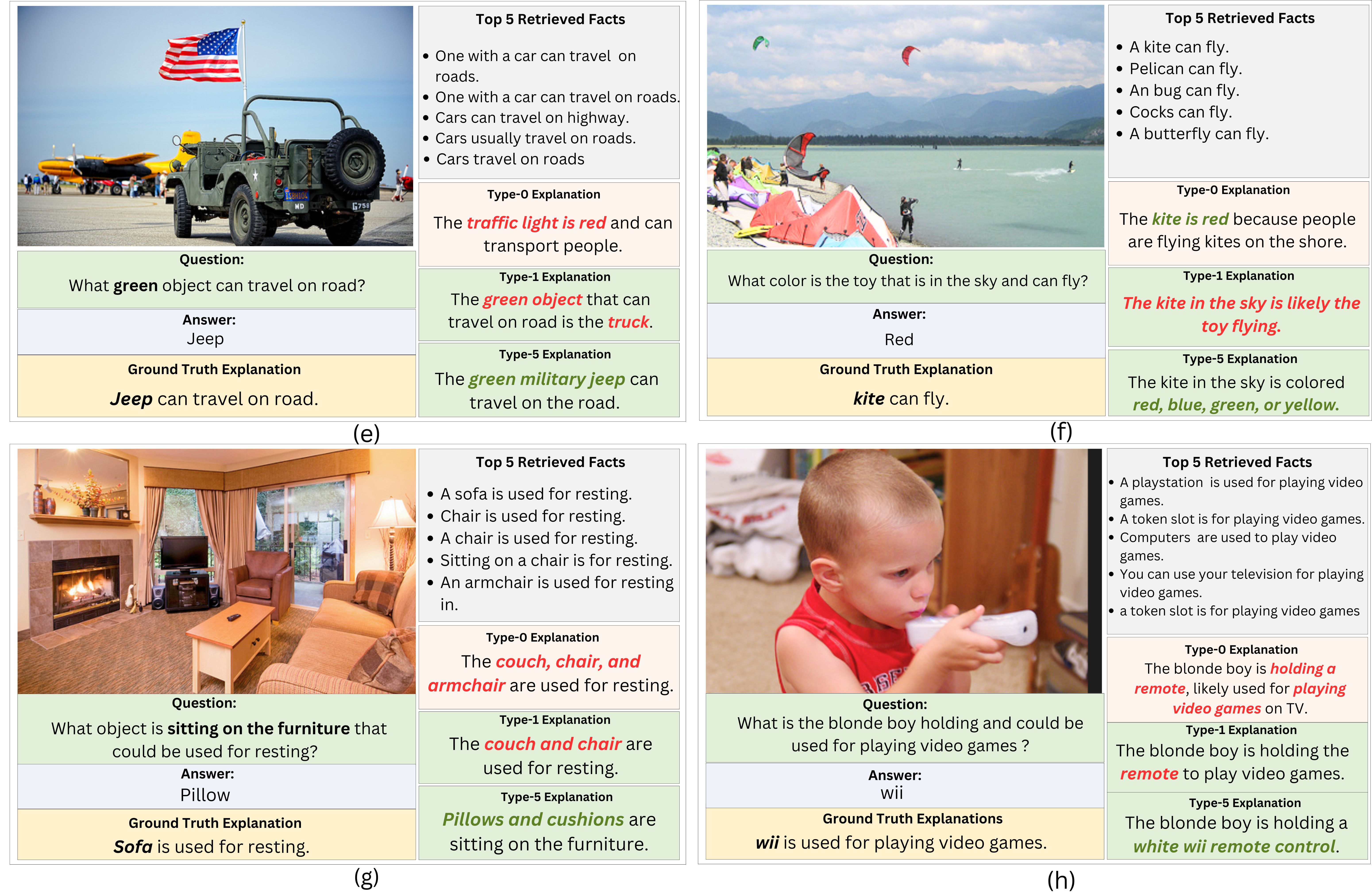} 
    \label{fig:explanation_collection3}    
\caption{More examples of \textbf{Type-0}, \textbf{Type-1}, and \textbf{Type-5} generated explanations.}
\end{figure*}

To avoid label leakage in the generated explanations, we also explicitly instruct the LLM not to produce the forbidden words. For instance, terms like "image description" and "caption" confuse the VQA model; thus, we filter out these words. This also includes words like "entailment" and "contradiction" when using the prompt to generate explanations for the e-SNLI-VE dataset. We explicitly instruct the LLM to generate only explanations within 10-15 words because, in the knowledge concatenation experiments, we found that the longer the input text gets, the lower the performance the model delivers.

\begin{table*}[t]
\small 
\centering
\label{tab:rouge_cumulative_scores}
\begin{tabularx}{\textwidth}{@{}ccccccccc@{}}
\toprule
Dataset & Knowledge Type & Split & Rougue-1 & Rougue-2 & \begin{tabular}[c]{@{}c@{}}Cumulative \\ 1-gram\end{tabular} & \begin{tabular}[c]{@{}c@{}}Cumulative \\ 2-gram\end{tabular} & \begin{tabular}[c]{@{}c@{}}Cumulative \\ 3-gram\end{tabular} & Cosine Score \\ 
\midrule
\multirow{6}{*}{e-SNLI-VE} & Explanations & Train & 24.35 & 5.97 & 13.20 & 6.31 & 3.88 & \textbf{49.98} \\
 & Retrieved Fact & Train & \textbf{27.11} & \textbf{8.06} & \textbf{20.92} & \textbf{10.58} & \textbf{6.62} & 47.66 \\
 \cmidrule(l){2-9} 
 & Explanations & Val & 25.79 & 6.43 & 14.50 & 6.85 & 4.20 & \textbf{50.36} \\
 & Retrieved Fact & Val & \textbf{31.87} & \textbf{7.81} & \textbf{20.17} & \textbf{10.04} & \textbf{6.26} & 47.46 \\
 \cmidrule(l){2-9} 
 & Explanations & Test & 25.95 & 6.54 & 14.50 & 6.85 & 4.17 & \textbf{50.50} \\
 & Retrieved Fact & Test & \textbf{32.08} & \textbf{7.98} & \textbf{20.30} & \textbf{10.14} & \textbf{6.34} & 47.04 \\
 \midrule
\multirow{6}{*}{CRIC} & Explanations & Train & 31.76 & 13.78 & 7.16 & 4.86 & 3.70 & 54.40 \\
 & Retrieved Fact & Train & \textbf{80.34} & \textbf{69.99} & \textbf{75.52} & \textbf{69.57} & \textbf{63.16} & \textbf{80.33} \\
 \cmidrule(l){2-9} 
 & Explanations & Val & 29.73 & 12.49 & 6.27 & 4.15 & 3.14 & 51.59 \\
 & Retrieved Fact & Val & \textbf{80.85} & \textbf{70.34} & \textbf{75.99} & \textbf{69.95} & \textbf{63.52} & \textbf{81.37} \\
 \cmidrule(l){2-9} 
 & Explanations & Test & 29.77 & 12.45 & 6.32 & 4.19 & 3.17 & 51.64 \\
 & Retrieved Fact & Test & \textbf{80.95} & \textbf{70.51} & \textbf{76.06} & \textbf{70.05} & \textbf{63.64} & \textbf{81.49} \\ 
 \midrule
\multirow{4}{*}{AOKVQA} & Explanations & Train & \textbf{49.29} & \textbf{27.86} & \textbf{39.88} & \textbf{29.55} & \textbf{23.45} & \textbf{65.31} \\
 & Retrieved Fact & Train & 0.3263 & 8.05 & 19.47 & 9.59 & \textbf{5.96} & 43.86 \\
 \cmidrule(l){2-9} 
 & Explanations & Val & \textbf{50.78} & \textbf{30.24} & \textbf{41.25} & \textbf{31.24} & \textbf{25.43} & \textbf{66.52} \\
 & Retrieved Fact & Val & 33.56 & 8.81 & 20.32 & 10.15 & 6.27 & 45.09 \\
 \bottomrule
\end{tabularx}
\caption{LLM-generated explanations outperform the retrieved facts in assisting the models in deriving correct answers due to their relevant context, precision, and coherence. However, the retrieved facts, while being less helpful for reasoning, yield higher BLEU \& ROUGE scores than the generated explanations. This is attributed to the nature of BLEU and ROUGE metrics-- retrieved facts have more overlapping $n$-grams with the ground truth than the generated explanations, even if they lack contextual relevance and depth.}
\end{table*}

\paragraph{What We Define as Hallucinations and Their Constituents}
\label{sec:hallucination-description}
In our work, we used the term \emph{“Hallucination”} repeatedly to indicate any content in an LLM-generated explanation that is not supported by and in some cases directly contradicts the information in the question, the image, or any retrieved knowledge provided as context. In other words, whenever the generated text introduces details (objects, events, attributes, etc.) that do not appear in the image or are not logically inferable from the question and any external knowledge, we categorise those extraneous or fabricated details as hallucinations.

A typical example would be in Fig. \ref{fig:hallucinated_explanation_collection}d when the explanation states that the image contains \textit{“apples”} even though the visual data and object detections show \textit{"no apples"}. This also includes instances where the explanation confidently describes relationships or properties that are not depicted (e.g., \textit{“the dog is wearing a red collar”} when no collar is visible). These are \textbf{hallucinated} elements because they are neither prompted by the data nor verifiable through the available context.

\paragraph{Generating Type-5 Explanations with Various Llama Versions}
\label{sec:llama-variants}
Since \textbf{Type-5 }explanations are the ones that worked best for our use case, as seen in Table \ref{tab:explanation_quality_scores} and Table \ref{tab:main_accuracy_table}, we tried generating them with smaller models from the Llama family, aside from our primary Llama-3.1-8B model. We used Llama-3.2-1B \& 3B for this. We report the scores in Table ~\ref{tab:llama-variant-table}.

\begin{table}[!ht]
\centering
\resizebox{\columnwidth}{!}{%
\begin{tabular}{@{}ccccccc@{}}
\toprule
\textbf{Ver} & \textbf{B-1} & \textbf{B-2} & \textbf{R-1} & \textbf{R-2} & \textbf{R-L} & \textbf{C}  \\ 
\midrule
8B  & \textbf{31.08} & \textbf{22.77} & \textbf{46.30} & \textbf{24.63} & \textbf{42.78} & \textbf{63.08} \\
3B  & 27.34 & 19.96 & 44.54 & 23.31 & 41.40 & 61.03 \\
1B  & 23.10 & 16.76 & 39.60 & 20.07 & 35.60 & 55.87 \\
\bottomrule
\end{tabular}%
}
\caption{Quality of Type-5 explanations generated with different Llama versions.}
\label{tab:llama-variant-table}
\end{table}

\paragraph{AOKVQA Scores for Various Type-K Explanations}
\label{sec:explanation-scores-aokvqa}
We initially reported that deriving the Type-5 explanation was most beneficial to our CRIC dataset task. Table ~\ref{tab:explanation-scores-aokvqa-table} reports the same score for AOKVQA, indicating that Type-5 explanations perform the best irrespective of the dataset, both in terms of capturing the visual context or guiding the VQA models for better predictions.
\begin{table*}[!htb]
\small
\centering
\begin{tabular}{ccccccccc}
\toprule
\textbf{Type} & \textbf{Explanation} & \textbf{B-1} & \textbf{B-2} & \textbf{B-3} & \textbf{R-1} & \textbf{R-2} & \textbf{R-L} & \textbf{Cosine} \\ 
\midrule
\textbf{Type-2} & DC + Q + $F_{I,Q}$ & 40.38 & 29.92 & 23.92 & 49.38 & 28.16 & 44.88 & 65.98 \\
\textbf{Type-3} & RC + Q + $F_{I,Q}$ & 34.76 & 25.94 & 20.66 & 43.27 & 24.50 & 40.07 & 50.88 \\
\textbf{Type-4} & RC + O + Q + $F_{I,Q}$ & 38.09 & 27.86 & 21.99 & 46.55 & 25.50 & 42.56 & 59.93 \\
\textbf{Type-5} & DC + RC + O + Q + $F_{I,Q}$ & \textbf{41.15} & \textbf{31.34} & \textbf{25.43} & \textbf{50.78} & \textbf{30.24} & \textbf{46.54} & \textbf{66.52} \\
\textbf{Type-6} & DC + RC + O + Q & 39.13 & 28.84 & 22.96 & 48.52 & 27.23 & 43.90 & 64.95 \\
\textbf{Type-7} & I + $<TP>$ & 17.49 & 7.51 & 4.48 & 20.45 & 2.57 & 16.78 & 28.08 \\
\bottomrule
\end{tabular}
\caption{Results from the val split of AOKVQA showing a comparison of BLEU, ROUGE, and Cosine Scores (w.r.t. the GT explanation) for various types of explanations.}
\label{tab:explanation-scores-aokvqa-table}
\end{table*}

\section{Training Parameters and Resources}
We report the learning rate and number of epochs below for the models we have used for VQA and Retrieval tasks from the knowledge bases. We have used a dual NVIDIA-A40 GPU system, each with 46 GB of memory, for all the experiments.
\begin{table}[!ht]
\resizebox{\columnwidth}{!}{%
\begin{tabular}{@{}ccccc@{}}
\toprule
Architecture & LR & Batch Size & \#Epoch & \#Hours \\ 
\midrule
ViLT & 5e-5 & 32 & 6 & 15\\
VisualBERT & 4e-5 & 32 & 6 & 14\\
FLAVA & 4e-5 & 32 & 6 & 14.5 \\
ColBERTv2 & 1e-4 & 32 & 4 & 3\\ 
\bottomrule
\end{tabular}
}
\caption{Hyperparameters for Finetuning tasks}
\end{table}

\begin{tcolorbox}[colback=gray!5!white,colframe=red!75!black]
\label{prompt}
Given an image description, the task is to generate an explanation based on the following information.\\

Traditional-Caption-(TC): \texttt{traditionalcaption}$_I$.\\
Dense Caption (DC): \texttt{densecaption}$_I$.\\
Region Caption (RC): \texttt{regioncaption}$_I$.\\
Objects (O): \texttt{objects}$_I$.\\
Question (Q): $Q$.\\
Retrieved Facts (RF): $F_{I,Q}$.\\

You strictly need to produce a 15-20 word single-line explanation to help VQA models derive conclusions and nothing else.  
Forbidden words: **"image description", "captions"**.
\end{tcolorbox}

\end{document}